\documentclass{article}
\usepackage{arxiv}

\usepackage{amsmath}
\usepackage{amsthm}
\usepackage{amssymb}
\usepackage{graphicx}
\usepackage{booktabs}
\usepackage{longtable}
\usepackage{array}
\usepackage{geometry}
\usepackage{caption}
\usepackage{ltablex}  % For better longtable + text wrapping support
\usepackage{adjustbox}  % For resizing if needed

\title{Evaluating LLM Metrics Through Real-World Capabilities}

\author{
 Justin K. Miller \\
  School of Physics\\
  University of Sydney\\
  Camperdown, NSW 2006 \\
  \texttt{justin.k.miller@sydney.edu.au} \\
   \And
 Wenjia Tang \\
  School of Art, Communication and English\\
  University of Sydney\\
  Camperdown, NSW 2006 \\
  \texttt{wenjia.tang@sydney.edu.au} \\
}

\begin{document}
\maketitle

%%%% Abstract text to be placed here %%%%%%%%%%%%   

%Mention data being used and the results are specific to that data
\begin{abstract} 
As generative AI becomes increasingly embedded in everyday workflows, it is important to evaluate its performance in ways that reflect real-world usage rather than abstract notions of intelligence. Unlike many existing benchmarks that assess general intelligence, our approach focuses on real-world utility, evaluating how well models support users in everyday tasks. While current benchmarks emphasise code generation or factual recall, users rely on AI for a much broader range of activities—from writing assistance and summarization to citation formatting and stylistic feedback. In this paper, we analyse large-scale survey data and usage logs to identify six core capabilities that represent how people commonly use Large Language Models (LLMs): Summarization, Technical Assistance, Reviewing Work, Data Structuring, Generation, and Information Retrieval. We then assess the extent to which existing benchmarks cover these capabilities, revealing significant gaps in coverage, efficiency measurement, and interpretability. Drawing on this analysis, we use human-centred criteria to identify gaps in how well current benchmarks reflect common usage that is grounded in five practical criteria: coherence, accuracy, clarity, relevance, and efficiency. For four of the six capabilities, we identify the benchmarks that best align with real-world tasks and use them to compare leading models. We find that Google Gemini outperforms other models—including OpenAI’s GPT, xAI’s Grok, Meta’s LLaMA, Anthropic’s Claude, DeepSeek, and Qwen from Alibaba—on these utility-focused metrics.

\end{abstract}

\section{Introduction}

DeepSeek-R1, released in January 2025, was widely promoted as the leading AI model \cite{deepseek_release,bbc_2025,guardian_weekly,cnbc_2025,yahoo_2025}. Yet, such claims often relied on vague assertions and selective quotes rather than clear evidence. Subsequent AI models have similarly been declared superior, supported mainly by benchmarks centred around tasks like coding proficiency \cite{deepseek_coder_v2}, multilingual abilities \cite{multilingual_tokenizer}, and ancient script translation \cite{humanitys_last_exam}. Concerns have arisen regarding these benchmarks' practical relevance \cite{abdin2024phi4}, statistical robustness \cite{miller2024errorbars}, susceptibility to adversarial inputs \cite{he2023blindspots}, and reliance on superficial fluency over factual accuracy \cite{wang2024faireval}. Closed-source evaluations further exacerbate transparency issues \cite{balloccu-etal-2024-leak, gallifant2024gpt4review}, and automated metrics may penalise useful outputs for deviating from rigid reference answers \cite{sottana2023metrics, calais2024beyondaccuracy}.

This paper investigates how generative text-based AI is actually utilised in practice, identifying core capabilities that are critical in real-world interactions. While models are increasingly trained to align with human preferences: emphasizing helpfulness, harmlessness, and truthfulness via methods like reinforcement learning from human feedback (RLHF) and Constitutional AI \cite{ouyang2022training,bai2022constitutional}. However, they continue to be evaluated on benchmarks such as MMLU (a suite of academic multiple-choice tasks across diverse subjects) and AIME (a benchmark based on American math competition problems), which prioritise abstract problem-solving and cognitive performance \cite{wang2024mmlu-pro, vals2025aime}. This evaluation paradigm risks conflating intelligence with alignment, despite longstanding philosophical recognition that the ability to achieve goals does not guarantee ethical or value-aligned behaviour \cite{gabriel2020alignment}.

We evaluate whether existing benchmarks effectively capture these real-world interactions and propose a capability-aligned framework grounded in human-centred evaluation criteria. This approach aims to provide researchers, developers, and policymakers a clearer basis for assessing and comparing AI models according to practical effectiveness.

Generative AI produces content based on learned distributions from a vast multitude of different data sources \cite{feuerriegel2023generative, Vaswani2017, Raffel2020, poltronieri2019technical}. Text-to-text models, especially Transformer-based architectures like GPT-4 \cite{OpenAI2023}, LLaMA 3 \cite{Touvron2024}, and DeepSeek V3 \cite{DeepSeek2024}, dominate current applications. These models power widely adopted tools such as ChatGPT and Gemini, supporting tasks like writing and summarization \cite{cazzaniga2024gen, laskar2024systematic}. Their conversational capabilities arise from exposure to extensive linguistic data, enabling contextual adaptability \cite{wahde2022conversational, roumeliotis2023chatgpt}. Given their widespread use and established evaluation ecosystem, this paper focuses specifically on these text-to-text models.

\subsection{Generative AI Limitations}

Generative AI faces several notable limitations. Stability remains a significant issue, particularly in tasks such as automated assessments, where subtle changes can cause inconsistencies in results \cite{laskar2024systematic, dam2024complete}. Ethical concerns also persist, including biases, inappropriate content Generation, privacy risks, and misuse \cite{fui2023generative}. Tools like ChatGPT and Midjourney have highlighted AI's struggles with sensitive content and biases \cite{akpan2025conversational, JasonHu_RaceGenderBiasMidjourney}. Accuracy further complicates AI deployment, as Large Language Models (LLMs) can fabricate plausible but incorrect information, notably citations \cite{jeong2023study, algaba2025llmcitations}. Such persistent issues underscore the necessity for careful evaluation aligned with realistic user expectations and contexts.

\subsection{Patterns of AI Use and Implications for Benchmarking}

AI adoption has significantly expanded, with over 75\% of organizations deploying AI primarily to enhance operational efficiency and mitigate immediate risks \cite{mckinsey2025}. 88\% of AI users are non-technical employees, using generative AI tools predominantly for productivity-focused tasks such as writing, summarization, and idea Generation \cite{DeSmet2024HumanGenerativeAI}. This widespread adoption by non-specialists emphasizes that benchmarks for AI must prioritize ease of use, reliability, and relevance to common productivity workflows, rather than purely technical performance metrics.

Sector-specific variations further inform benchmark design. For instance, AI adoption is significant in legal (26\%), retail (40\%), and healthcare industries, the latter forecasting substantial growth in diagnostics and efficiency improvements \cite{EdgeDelta_AiAdoption}. These distinct usage contexts suggest that effective benchmarks must be versatile enough to accommodate industry-specific applications and performance standards.

Individual engagement with generative AI, such as ChatGPT (used by 23\% of U.S. adults), primarily involves text Generation for professional, educational, and entertainment purposes, a trend particularly strong among younger demographics \cite{pew2024}. This aligns closely with industry trends, highlighting that benchmarks should effectively evaluate AI's capabilities in realistic, text-oriented tasks reflective of actual usage.

The education sector provides a valuable example of the ethical complexities surrounding AI adoption, with significant numbers of students using generative tools despite acknowledging issues of academic integrity \cite{polyportis2024, intelligent2023}. Thus, benchmarks should not only measure technical proficiency but also consider transparency, explainability, and ethical guidelines as critical evaluation criteria.

In technical fields such as software development, AI integration is driven by tangible efficiency gains, with adoption rates reaching 62\% among developers who primarily value task acceleration and productivity improvements \cite{stackoverflow2024}. Benchmarks aimed at technical sectors should therefore explicitly measure productivity enhancement, integration ease, and reliability in code Generation and debugging tasks.

Finally, professional task alignment strongly influences AI adoption, with higher integration observed in fields like marketing and journalism, and cautious approaches in finance, healthcare, and education due to heightened concerns around accuracy, privacy, and ethical responsibility \cite{humlum2024}. This pattern suggests benchmarks must account for domain-specific tolerances for error and incorporate assessments that reflect real-world consequences of AI inaccuracies or misuse.

Collectively, these adoption trends highlight the necessity for human-centric AI benchmarks that not only capture technical capability but also meaningfully assess the practical value, ethical considerations, and contextual relevance of AI across diverse sectors and user groups.

\subsection{AI use as Conversation}

Text-to-text generative AI inherently adopts a conversational structure, characterized by user prompts and corresponding AI-generated responses \cite{mctear2022conversational}. This has frequently led to their framing as artificial intelligence-based chatbots, for example, in educational \cite{limna2023use}, medical \cite{rasmussen2023artificial}, and commercial contexts such as Google’s Gemini \cite{carla2024exploring}, Microsoft’s Copilot \cite{durach2024hello}, and META’s LLAMA \cite{wang2024faireval}. Such portrayals underscore dialogue's integral role in generative AI's functioning—interpreting instructions, engaging users, facilitating iterative task completion, and aligning closely with user intent \cite{fui2023generative, jeong2023study}.

Despite this conversational framing, existing benchmarks inadequately capture the complexities of genuine conversational interaction. Historically, evaluative frameworks like Grice's Cooperative Principle—with maxims emphasizing clear, sufficient, relevant, and contextually appropriate communication—have guided conceptions of effective human dialogue \cite{grice1975logic, grice1978further}. Such principles reflect essential conversational dimensions, including mutual understanding, cooperation, and grounding to prevent misunderstandings \cite{clark1991grounding, yang2023understanding}. However, this standard cannot be directly adopted into human-to-machine communication with different contexts and scenarios. Current AI evaluation standards predominantly focus on discrete task performance, neglecting critical conversational qualities such as iterative clarification, contextual relevance, and implicit meaning interpretation \cite{LevesqueGrice2021, goffman1974frame, sacks1974simplest}.

Moreover, human-computer dialogues exclusively rely on textual communication without supplementary non-verbal cues, which are natural qualities of conversation between humans \cite{Turkle2011}. Effective communication with AI thus demands explicit attention to context, subtext, and implied meanings—elements often absent from some prevailing metrics.

To bridge these gaps, this study proposes a capability-aligned, human-centred evaluation framework explicitly designed around conversational dynamics. This framework integrates key conversational dimensions—coherence, accuracy, clarity, relevance, and efficiency—and emphasizes iterative adaptation and learning \cite{berger2010handbook}. By aligning benchmarks with authentic conversational interactions, this approach aims to enhance AI's practical effectiveness in real-world applications.

\subsection{What are ``Good'' AI Conversations?}

Building upon historical research on conversational quality and contemporary expectations for generative AI, we draw from classic theories of human-to-human conversation and incorporate characteristics of AI usage contexts to propose criteria for systematically evaluating interactions between LLMs and human users. These criteria are divided into two clear categories: \textit{objective criteria}, which are directly observable and measurable within the conversation itself, and \textit{subjective criteria}, which reflect users' personal evaluations and perceptions regarding conversational effectiveness and task performance.

\subsubsection*{Objective Criteria}

Objective criteria encompass measurable conversational characteristics directly observable without requiring subjective user interpretation:

\begin{itemize}
    \item \textbf{Coherence:} Maintaining logical consistency and structured dialogue is essential for effective communication \cite{venkatesh2017evaluating}. Coherent conversations enable both human and AI participants to effectively integrate previous conversational context, clarify ambiguities, and address implicit meanings, thereby collaboratively progressing toward shared conversational objectives \cite{grice1978further}.

    \item \textbf{Accuracy:} Accuracy directly determines the reliability of information provided by AI, thus significantly impacting task effectiveness and user trust \cite{zhang2020effect}. Generative AI often produces inaccuracies or entirely fabricated information—a phenomenon known as ``\textit{hallucination}''—due to inherent limitations in adapting to new or unseen data \cite{euchner2023generative, jeong2023study}. Consequently, validating content accuracy and truthfulness is essential for practical scenarios where precise and trustworthy outcomes are necessary, such as medical, legal, or financial contexts \cite{laskar2024systematic}.

    \item \textbf{Relevance:} Providing relevant information focused explicitly on the task significantly improves conversational effectiveness. Extraneous or redundant details can obscure intent, disrupt user focus, and diminish the quality and effectiveness of the interaction \cite{grice1975logic}. Therefore, AI responses should prioritize delivering content directly aligned with achieving user-defined objectives.

    \item \textbf{Clarity:} Effective conversations prioritize concise and clear expressions, thereby reducing cognitive demands on users. By avoiding unnecessary complexity, users can quickly interpret and utilize AI-generated content without additional cognitive overhead or extensive clarification \cite{venkatesh2017evaluating}.

    \item \textbf{Efficiency:} Conversational efficiency involves proactively providing and requesting only essential information required for task completion, reducing redundancy, and minimizing cognitive overload \cite{gerlich2025ai}. Efficient interactions optimize the conversational process, ensuring streamlined task accomplishment without unnecessary cognitive or temporal expenditures.
\end{itemize}

\subsubsection*{Subjective Criteria}

Subjective criteria involve user perceptions and emotional responses, significantly influencing the perceived quality and task-oriented success of interactions:

\begin{itemize}
    \item \textbf{Politeness:} Politeness significantly impacts users’ perceived satisfaction and conversational quality. While users recognize AI as non-human entities, polite interactions foster social presence, enhance user comfort, and encourage user engagement \cite{reeves1996media}. Politeness thus indirectly supports effective task completion by increasing users' willingness to cooperate, provide thorough input, and positively engage with AI tools.

    \item \textbf{Conversation Structure:} The conversational structure—encompassing vocabulary selection, linguistic style, and adherence to conversational norms—directly influences perceived effectiveness and user satisfaction \cite{bansal2024transforming, guydish2021good, ishii2019revisiting}. Adhering to established social and conversational norms enhances perceived interaction quality, indirectly improving task outcomes by fostering user comfort and engagement \cite{chen2024feasibility}.

    \item \textbf{Satisfaction:} User satisfaction reflects the alignment between conversational outcomes and user expectations. Unlike purely transactional interactions, users typically engage AI tools seeking supportive, reliable assistance analogous to interactions with peers or friends \cite{gupta2023practical}. Satisfaction directly impacts user willingness to repeatedly utilize AI tools and ultimately affects perceptions of conversational effectiveness and success in task-related contexts.

    \item \textbf{Motivation:} Addressing intrinsic motivational needs—including emotional fulfillment, personal security, and recognition—improves users' emotional investment and trust in conversational exchanges \cite{yang2023understanding}. AI tools capable of recognizing and addressing these motivational dimensions enhance overall user engagement, indirectly improving task performance by sustaining user interest, cooperation, and trust.
\end{itemize}

These criteria, synthesized from prior literature and recent AI advancements, provide clear theoretical benchmarks for conversational quality. Understanding these expectations enables a structured evaluation of real-world AI performance, identifying demonstrated capabilities and highlighting areas for targeted improvements. The following sections explore how well current AI implementations align with these established conversational expectations, clarifying the existing gaps and opportunities for future enhancements.

\section{Methods}

To investigate how AI is being used in general tasks, we built on two key sources: a large-scale survey of Danish workers \cite{humlum2024}, and an empirical analysis of generative AI usage using Anthropic’s Claude.ai platform \cite{handa2025economic}.

\subsection{Identifying AI-Capable Tasks}
To identify the types of occupational tasks most amenable to AI assistance, we conducted a small-scale exploratory analysis using data from a study that surveyed over 18,000 workers across 11 occupations in Denmark and linked survey responses with administrative labor market records \cite{humlum2024}. The authors used a refined version of OpenAI’s “Direct Exposure (E1)” measure to assess whether GPT-3.5 could significantly reduce the time required to complete specific job tasks. A task was marked as providing “Large” time savings if access to ChatGPT could halve the time needed for an average worker to complete it, at equivalent quality.

These task ratings were generated through a hybrid process involving GPT-3.5 prompting and human validation. Specifically, the authors assessed Detailed Work Activities (DWAs) from the U.S. O*NET database, selecting six representative tasks per occupation based on their average exposure scores and importance. Of the 66 tasks, the final sample included 21 tasks across the 11 occupations, all rated as having high productivity potential through ChatGPT assistance \cite[SI Appendix, Section 1.A]{humlum2024}.

From this list of high-exposure tasks, we performed qualitative thematic analysis \cite{braun2006} to categorize the types of tasks where ChatGPT offers the greatest productivity gains. Each author independently reviewed the 32 tasks and grouped them by thematic similarity based on the cognitive processes and outputs involved. We iteratively discussed and refined these groupings until reaching consensus on six recurring AI capabilities. These categories are not mutually exclusive; individual tasks were often associated with multiple capabilities reflecting the overlapping nature of real-world work.

See Table \ref{tab:ai-capabilities-tasks} for a complete mapping of each task to its assigned capabilities. This preliminary classification serves as the basis for developing capability-aligned AI evaluation metrics.

\subsection{Validating Capabilities with Real-World Usage}

To test the robustness of the AI capability categories found during the qualitative thematic analysis, we validated them using a second data source \cite{handa2025economic}. This source mapped over four million Claude.ai prompts to O*NET tasks and grouped them by occupational categories.

We accessed the dataset provided by Anthropic, extracting the specific O*NET tasks, their associated occupational categories, and the percentage of total prompts linked to each task. In total, the dataset included 35,014 unique tasks. A cumulative distribution plot of task frequency is shown in Figure~\ref{fig:ai_cumulative}. Notably, the top 100 most frequent tasks accounted for just over 50\% of all usage. 

We therefore focused our analysis on these top 100 tasks, which represent over two million real-world AI prompts. This approach enabled a broad coverage of usage while keeping the qualitative review tractable. Each task was manually reviewed by the authors and mapped to one or more of the six AI capabilities identified earlier.

\subsection{Constructing Prompts and Usage Mapping}

For each of the top 100 tasks, we assigned at least one AI capability, with many tasks mapping to multiple capabilities. To illustrate how each capability manifests in practice, we constructed an example prompt for every task–capability pairing. These prompts were designed to reflect how a user might realistically request assistance from a language model when performing the given task.

Creating example prompts served two purposes. First, it ensured that the mapping between tasks and capabilities was grounded in real-world usage, not just abstract categorisation. By imagining a plausible user prompt, we were able to validate whether a capability truly applied to the task and how it might be operationalized in a typical human–AI interaction. Second, the prompts provided an interpretable bridge between occupational tasks and the kinds of language-based inputs that LLMs are designed to handle. This step enhanced the interpretability of our mapping and ensured that our subsequent evaluation remained anchored in naturalistic language use. The full table for the top 100 tasks and example prompts given by the authors for each task can be found in Table \ref{tab:top100-anthropic-tasks}

Using the usage percentage data provided by Anthropic, we then computed the cumulative usage share for each AI capability. This allowed us to quantify how heavily each capability featured in real-world AI interactions. The resulting distribution is shown as a comparative bar chart in Figure~\ref{fig:ai_cap}, highlighting the relative importance of each capability based on actual usage frequency.

\subsection{Evaluation of Metrics}

To assess how well existing AI benchmarks reflect real-world usage, we evaluated a set of widely used metrics through the lens of the six AI capabilities in Appendix B Table \ref{tab:ai_capabilities_extended}. This step completes the methodological framework by mapping current benchmarks to the practical functions users expect from language models.

We based our evaluation on five objective qualities commonly expected in human–AI interactions: coherence, accuracy, clarity, relevance, and efficiency. These criteria serve as a human-centered lens for assessing whether a benchmark realistically captures how users interact with AI systems. Each criterion is defined below

\begin{itemize}
    \item \textbf{Coherence:} Does the metric assess AI performance using prompts and formats that resemble genuine human interaction, rather than artificial test structures, such as multiple choice?
    
    \item \textbf{Accuracy:} Is the information verifiably correct according to trusted sources or gold-standard answers? We examined whether each metric had a clear standard for correctness and the types of sources it referenced.
    
    \item \textbf{Clarity:} Does the metric measure whether outputs are easy to understand and clearly worded? Even a technically correct output may be unhelpful if it is confusing or overly complex.
    
    \item \textbf{Relevance:} Does the benchmark test a broad and meaningful range of content within the capability domain? For instance, a programming benchmark focused solely on python would not necessarily help with coding problems in Java.
    
    \item \textbf{Efficiency:} Does the metric reflect how much time or cognitive effort the AI helps save? While often overlooked, this is critical to assessing practical utility in workplace settings.
\end{itemize}

We reviewed the release notes and technical specifications of recent foundation models from OpenAI \cite{openai_gpt45}, Google Gemini \cite{google2025gemini2.5}, Deepseek \cite{deepseek2025r1}, xAI \cite{xai2025grok3}, Qwen \cite{qwenlm2025qwen2.5max}, Meta \cite{meta2025llama4}, and Anthropic \cite{anthropic2025claude37sonnet}. From these sources, we compiled a list of benchmarks commonly used to evaluate text-to-text generative capabilities:

\begin{quote}
\textit{MMLU, AIME, Codeforces, GPQA, SWE-bench Verified, FACTS Grounding, MRCR, HumanEval, SimpleQA, GSM8K, Math-500, Bird-SQL, LiveCodeBench, Humanities Last Exam, MMLU-Pro, AGIEval English, CRUXEval-I/O, SimpleQA, Chatbot Arena, Webdev Arena} \cite{hendrycks2020mmlu,vals2025aime,quan2025codeelo,rein2023gpqa,princeton2025swebench,jacovi2025facts,vodrahalli2024michelangelo,openai2021humaneval,wei2024simpleqa,cobbe2021gsm8k,yao2024tau,vals2025math500,li2023birdsql,livecodebench2025,phan2025hle,wang2024mmlu-pro,cui2025agieval,gu2024cruxeval,wu2025writingbench,vichare2025webdev,chiang2024chatbot}
\end{quote}

Each metric was then mapped to the AI capability it most directly tests. To assess alignment, we reviewed the benchmark’s design, inputs, outputs, and scoring methodology to determine how well it addressed each of the five human-centered evaluation criteria: coherence, accuracy, clarity, relevance, and efficiency. This set of five criteria constitutes our evaluation framework. 

For each capability, we then selected the benchmark that best aligns with all five dimensions, based on our independent reviews of the benchmark documentation and, where applicable, third-party validation studies. Full assessments for each benchmark are included Appendix A Table~\ref{tab:appendix-metric-eval}.

\section{Results}

\begin{figure}[ht]
    \centering
    \includegraphics[width=0.7\textwidth]{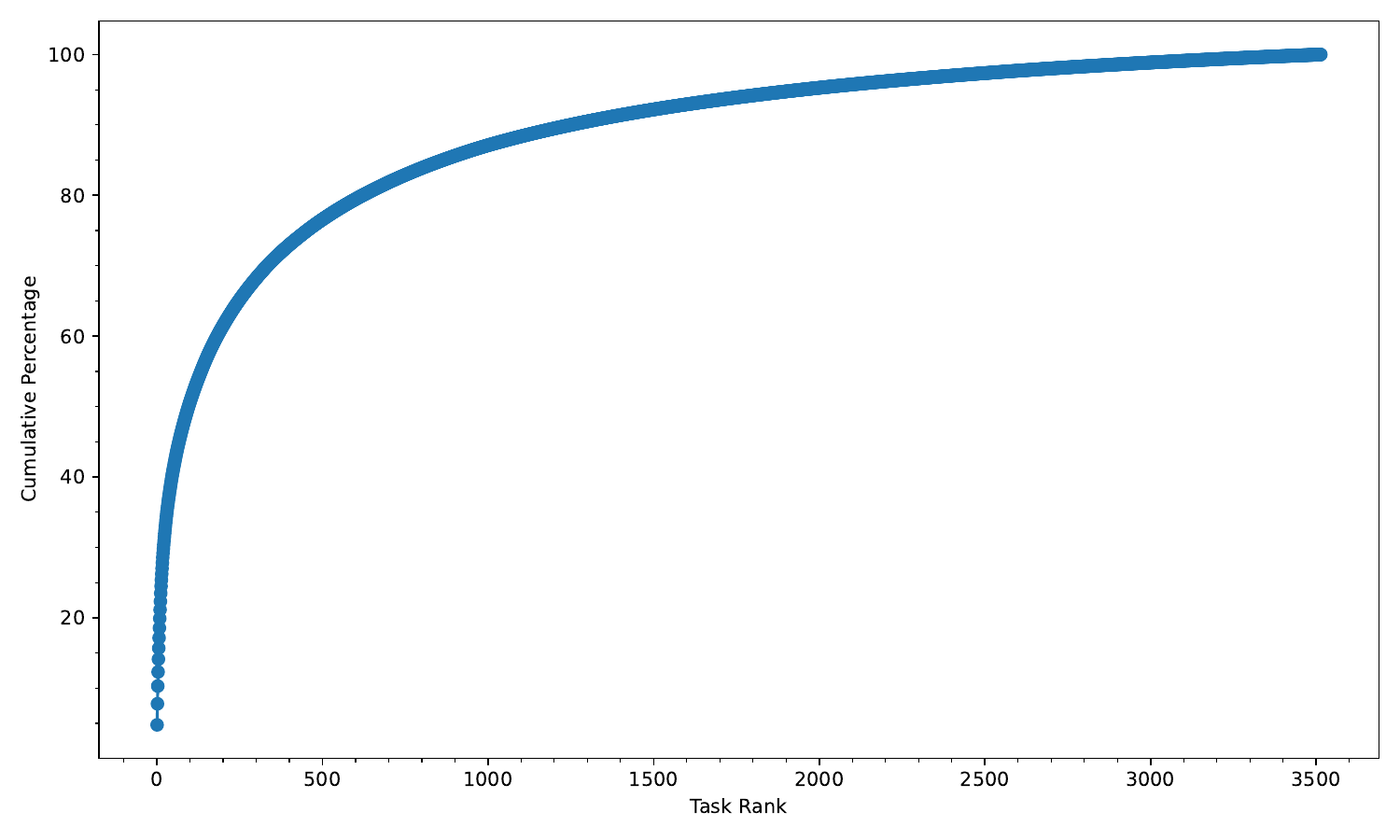}
    \caption{Cumulative distribution of AI usage across 35,014 occupational tasks. Each task is ranked by the percentage of total Claude.ai prompts associated with it, using data from Handa et al. (2025) \cite{handa2025economic}. The x-axis shows task rank in descending order of frequency, and the y-axis represents the cumulative percentage of total prompts. A small number of tasks dominate usage: the top 100 tasks account for just over 50\% of all prompts.}
    \label{fig:ai_cumulative}
\end{figure}

In Figure \ref{fig:ai_cumulative}, we observe a cumulative distribution curve illustrating how AI usage is concentrated across 35,014 occupational tasks, based on Claude.ai prompt data from Handa et al. (2025) \cite{handa2025economic}. The x-axis ranks tasks from most to least frequently prompted, while the y-axis shows the cumulative percentage of total prompts.

The steep initial rise of the curve reveals a strong skew: a small subset of tasks accounts for a disproportionately large share of usage. Notably, the top 100 tasks alone comprise just over 50\% of all prompts, while the top 500 account for just under 80\%. This long-tail distribution indicates that although AI is applied across a broad range of tasks, actual usage is highly concentrated in a narrow band of frequently occurring requests.

\subsection{AI Capabilities}

In Table \ref{tab:ai_capabilities_extended}, we present six core AI capabilities that emerged from thematic analysis of occupational tasks marked as highly automatable, as well as validation using real-world usage data from Claude.ai. These capabilities: Summarization, Technical Assistance, Reviewing Work, Data Structuring, Generation, and Information Retrieval, capture the main functions AI performs to support workers across diverse domains. Each reflects a distinct mode of interaction: for example, Information Retrieval supports fact-finding without producing new and original content, while Generation involves the creation of novel material. The examples included illustrate how the same capability can span both objective and subjective domains. Summarization, for instance, may involve extracting price trends from datasets (objectively measurable) or summarizing an author’s viewpoint (subjective).

The table also distinguishes between objective and subjective types of evaluation. A key distinction between these two types is that objective metrics can often times be automated, and thus are easier for metrics to measure, while Subjective measurements require human evaluation. These distinctions offer insight into where AI performance can be reliably benchmarked and where human input remains essential. We validated the robustness of these categories using the top 100 most frequently used occupational tasks which account for over 50\% of four million Claude.ai prompts. Each of the top 100 most frequently used tasks could be linked to at least one of the six capabilities, and many to multiple.

\begin{table}[ht!]
\centering
\begin{tabular}{|p{3cm}|p{5cm}|p{2cm}|p{5cm}|}
\hline
\textbf{AI Capability} & \textbf{Explanation} & \textbf{Type of Evaluation} & \textbf{Examples} \\
\hline
\textbf{Summarization} & Analysing large amounts of content to extract key information and present concise summaries & Subjective &  Summarise the author’s main arguments in the web pages \cite{GoogleAssistant_2023} \\
\hline
\textbf{Technical Assistance} & Providing clear instructions, diagnosing issues, and suggesting step-by-step solutions for software and hardware problems & Objective & Write optimized code \cite{busu2024using} \\
\hline
\textbf{Reviewing Work} & Identifying issues, evaluating systems or processes, and recommending improvements or solutions & Subjective and Objective & Objective correct math homework answers \cite{Rogers2024} Subjective: review my email for tone and clarity and suggest improvements \\
\hline
\textbf{Data Structuring} & Efficiently managing, logging, and maintaining accurate records of transactions, interactions, or documentation & Objective &Change academic reference list from Chicago to APA style \cite{khalifa2024using}\\
\hline
\textbf{Generation} & Creating original ideas, drafting written content, or suggesting creative solutions based on provided parameters or goals & Subjective & Create content for an About Us page for a small bakery.\\
\hline
\textbf{Information Retrieval} & Finding and delivering relevant factual or background information in response to queries, often based on  pre-trained knowledge or Retrieval-augmented Generation (RAG) & Objective & What is the current price of mangoes in Australia \\
\hline
\end{tabular}
\caption{This table presents a multidimensional framework connecting six real-world AI capabilities with key evaluation criteria. Each capability (e.g., Summarization, Generation) is assessed using both objective and subjective criteria—reflecting whether the capability can be evaluated is based on observable outputs (objective) or user perception (subjective). This structure highlights how current benchmarks align with, or overlook, practical AI uses by showing which criteria are essential for evaluating each capability. The intersecting dimensions are necessary to capture the full complexity of AI performance in real-world tasks.}
\label{tab:ai_capabilities_extended}
\end{table}

\begin{figure}[ht]
    \centering
    \includegraphics[width=0.7\textwidth]{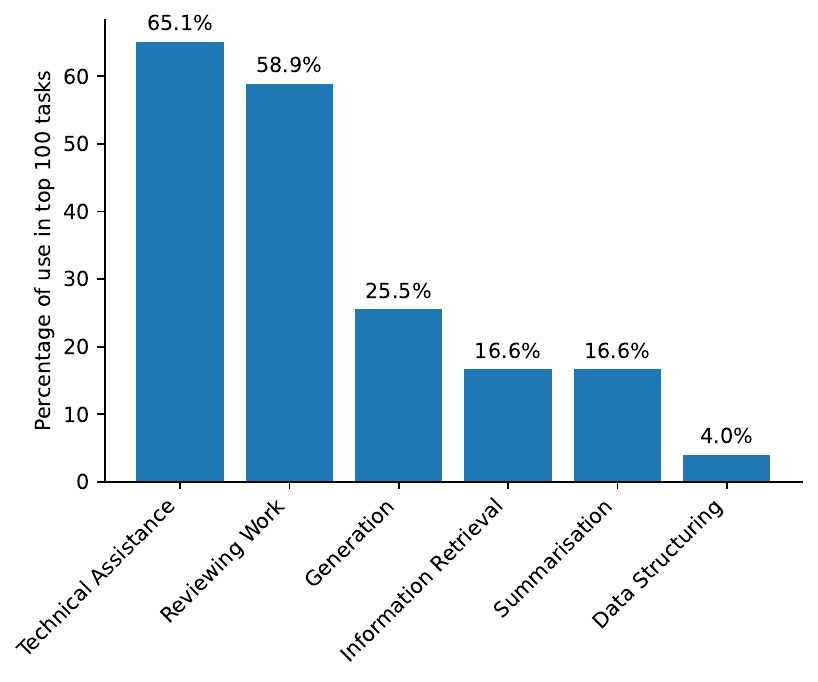}
    \caption{Percentage of the top 100 occupational tasks for which each AI capability is relevant. A task can be associated with multiple capabilities. For example, the task \textit{“modify existing software to correct errors, to adapt it to new hardware, or to upgrade interfaces and improve performance”} involves both Technical Assistance and Reviewing Work. The percentages reflect the sum of the percentage scores (derived from over 4 million prompts) associated with each task–capability pair, divided by the maximum cumulative importance across the top 100 tasks. This highlights how commonly each capability appears among the tasks people are most likely to use AI for.}
    \label{fig:ai_cap}
\end{figure}

Figure \ref{fig:ai_cap} shows the relative importance of each AI capability across the top 100 most commonly used occupational tasks, as identified in over four million real-world Claude.ai prompts. Each bar reflects the cumulative percentage of task–capability associations, normalized by the maximum cumulative importance observed. Importantly, tasks can be linked to more than one capability—for example, the task “modify existing software to correct errors, to adapt it to new hardware, or to upgrade interfaces and improve performance” requires both Technical Assistance and Reviewing Work. This approach captures not just how many tasks are associated with each capability, but how heavily those tasks feature in real-world AI use.

The results highlight Technical Assistance (65.1\%) and Reviewing Work (58.9\%) as the most prevalent capabilities, suggesting that AI is most often used to support problem-solving, diagnosis, and improvement-related activities. In contrast, capabilities such as Generation (25.5\%), Information Retrieval (16.6\%), and Summarization (16.6\%) are moderately represented, while Data Structuring (4.0\%) is relatively rare among the top-used tasks. These findings suggest that users most frequently rely on AI for tasks that require operational or evaluative reasoning, while tasks involving routine formatting play a less prominent role.

\subsection{Metrics Used by AI Companies}
\label{sec:metrics}
Of the six AI capabilities we identified, we found that existing benchmarks covered only three of them: \textit{Information Retrieval, Technical Assistance}, and \textit{Summarization}. For these three, we were able to identify relevant and widely-used metrics. For instance, MMLU and AGIEval English were most aligned with Information Retrieval, while SWE-bench and HumanEval were relevant to Technical Assistance.

To assess whether the remaining capabilities were covered by existing benchmarks, we conducted a structured search of academic literature, leaderboard repositories, and benchmark evaluations for newer foundation models. Our goal was to identify any widely used benchmarks that could be reasonably mapped to the remaining capabilities: Generation, Reviewing Work, and Data Structuring.

This search revealed WritingBench as a suitable candidate for evaluating the Generation capability \cite{wu2025writingbench}. WritingBench includes structured tasks across multiple writing domains—such as persuasive, technical, and informative writing—and uses a fine-tuned critic model validated against human judgments to evaluate output quality. These features align closely with the Generation capability as defined in our framework, which involves creating original content across diverse goals and genres.

In addition, we identified the Creative Writing section of Chatbot Arena as another relevant benchmark for this capability \cite{chiang2024chatbot}. It evaluates open-ended Generation tasks such as storytelling and poetry through pairwise human preference voting, providing a comparative score that reflects subjective judgments of human preference. Together, these two benchmarks represent the only benchmarks found that reflect the Generation capability.

For Reviewing Work and Data Structuring, however, we found no widely adopted benchmarks that explicitly test these capabilities. In the absence of suitable benchmarks, we did not attempt to assign scores to these capabilities. Instead, for the four capabilities where relevant benchmarks did exist—Information Retrieval, Technical Assistance, Summarization, and Generation—we collected publicly reported benchmark scores from leading models developed by Google, Anthropic, OpenAI, xAI, Meta, DeepSeek, and Alibaba. For each capability, we selected the highest-scoring model per company, using confidence intervals or reported variance where available to facilitate meaningful comparisons.

We now examine the extent to which existing benchmarks capture these core capabilities, identifying areas of alignment as well as major blind spots in practical utility.

\subsubsection{Technical Assistance}
Among the evaluated benchmarks for Technical Assistance, WebDev Arena stands out as the most realistic and user-centered according to our criteria \cite{vichare2025webdev}. In terms of coherence, WebDev Arena uses open-ended prompts that closely resemble how humans naturally ask for help in practical settings (e.g., “Build a chess game” or “Clone the WhatsApp UI”). Unlike benchmarks such as HumanEval or CRUXEval-I/O, which the authors of this paper examined and found to focus on narrowly defined tasks implemented as Python functions, WebDev Arena reflects real-world requests that are broader, more ambiguous, and closer to how LLMs are actually deployed. For accuracy, while WebDev Arena does not rely on gold-standard answers, it uses human preference voting as a judgment mechanism, allowing outputs to be evaluated in a context-sensitive manner. This contrasts with benchmarks like SWE-bench Verified, where nearly one-third of correct solutions involved cheating or weak test cases \cite{aleithan2024swebenchplus}, and Codeforces, which suffers from dataset leakage over time \cite{huang2023competition}. On the dimension of clarity, WebDev Arena does not explicitly penalize unclear output, but human preference may implicitly reflect whether responses are well-structured and comprehensible—an approach that is arguably more aligned with end-user needs than automated tests like pass@k used in HumanEval. Regarding relevance, WebDev Arena’s major strength is its focus on front-end web development using HTML, CSS, and JavaScript across eleven broad categories and many subcategories. In contrast, most other benchmarks—such as Codeforces, HumanEval, LiveCodeBench, CRUXEval-I/O, and SWE-bench Verified—evaluate tasks exclusively in Python. This makes WebDev Arena more representative of real-world UI and client-side development but also limits its relevance for back-end or systems-level programming. Finally, while none of the benchmarks measure efficiency directly, WebDev Arena indirectly reflects it through human preference, capturing how helpful or time-saving an LLM-generated solution feels to a user. By contrast, Codeforces is one of the few to report task completion time, and Bird-SQL introduces a reward-based efficiency score, though both rely on more narrowly scoped or language-specific tasks. Overall, WebDev Arena offers the most practical and human-aligned assessment of Technical Assistance, even while acknowledging its trade-offs in backend evaluation and objective scoring.

\subsubsection{Information Retrieval}

Among the benchmarks evaluated for Information Retrieval, SimpleQA emerges as the most well-rounded according to our criteria. It performs better than other benchmarks on coherence, as it presents concise, fact-based prompts and expects short-form answers, rather than relying on multiple-choice formats which diverge from how humans typically interact with language models. In contrast, benchmarks like MMLU, GPQA, and AGIEval all adopt rigid multiple-choice structures with fixed answer formats (for example, "The answer is A"), which limit natural interaction and reduce applicability in real-world retrieval settings. Accuracy in SimpleQA, while not flawless, has undergone independent review with 94.4\% agreement among expert annotators, providing a relatively high standard of validation\cite{wei2024simpleqa}. This is notably more robust than GPQA, which has only expert agreement of 74\% \cite{rein2023gpqa}, and GSM8K, which initially included a high proportion of incorrect questions before refinement in the Platinum version \cite{vendrow2025llmreliability}. MMLU has an overall accuracy of 6.49\%, however these errors are not distributed uniformly among topics with the Virology section having an error rate of 57\%. This error rate is an issue as most modern models are getting close to 90\% \cite{deepseek2025r1}. For clarity, SimpleQA implicitly encourages clear, succinct answers, though like most other benchmarks—including Humanities Last Exam and Math-500—it does not explicitly assess how understandable the outputs are for human users. On relevance, SimpleQA covers a broad and diverse range of knowledge domains such as STEM, politics, art, geography, sports, and more, setting it apart from domain-specific benchmarks like GSM8K, AIME, and Math-500, which are focused solely on mathematics, or GPQA, which is limited to science subjects. Even the broader MMLU and Humanities Last Exam, though more interdisciplinary, often reflect a US-centric or highly academic framing that may not translate well to everyday queries. Finally, although efficiency is not directly measured in SimpleQA, this is a limitation shared by nearly all benchmarks in this space—only AIME provides an explicit latency-to-answer measure. On balance, SimpleQA offers the clearest, most coherent, and domain-relevant framework for evaluating Information Retrieval capabilities in language models.

\subsubsection{Summarization}
Among the limited benchmarks available for evaluating the Summarization capability, MRCR provides a more rigorous and direct test of a model’s ability to condense and retrieve information from large contexts. Unlike FACTS Grounding, which relies on domain-specific materials such as legal texts, Wikipedia entries, and scientific documents, MRCR employs synthetic data that is carefully constructed to test core summarization skills—particularly the ability to distinguish and retrieve closely related pieces of information based on structure and context. This makes MRCR less reliant on background knowledge and more focused on the underlying cognitive capability being measured. While both benchmarks are comparable in coherence and clarity, FACTS Grounding introduces an epistemological issue by relying on an AI model from a different family to assess correctness. Although the designers take steps to avoid evaluation bias by not using models from the same family, the absence of a consistent comparison to human judgement raises concerns about the validity of its accuracy scores. MRCR, by contrast, provides more objective performance signals based on retrieval of precisely planted information. Neither benchmark explicitly measures efficiency, and both could be improved by incorporating user-centric metrics like time saved or effort reduced. Overall, despite the limited number of summarization benchmarks compared to other capabilities, MRCR offers a more principled and interpretable framework for evaluating summarization in language models.

\subsubsection{Generation}
Among the very limited benchmarks currently available for evaluating the Generation capability, Chatbot Arena: Creative Writing offers the most human-aligned and comparative approach. It uses head-to-head human preference voting to evaluate model responses to creative prompts, such as writing poems, short stories, or humorous narratives. This evaluation method aligns closely with how real users experience AI-generated content, making it particularly strong on coherence and clarity—two dimensions that are naturally captured through direct human judgment. Moreover, Chatbot Arena provides an aggregated score for each model, allowing comparisons across systems, which is valuable for tracking progress in open-ended Generation tasks. However, WritingBench also brings important strengths. Unlike the open-ended and sometimes subjective setup of Chatbot Arena, WritingBench is structured around clearly defined writing tasks spanning six core domains and one hundred subdomains, including persuasive, technical, and informative writing. It uses a fine-tuned critic model validated against human judgments to assess performance based on explicit criteria, including clarity, structure, and genre appropriateness. One key limitation shared by both benchmarks is the absence of an explicit measure of efficiency—neither evaluates how much cognitive effort or time is saved for the user. Nonetheless, given the field’s current landscape, Chatbot Arena stands out for its human-grounded evaluation method and breadth of comparative scoring, while WritingBench remains a promising complement for more structured and evaluable writing tasks. Overall, the lack of established benchmarks in this area highlights that content Generation is an under-researched area within AI capabilities.

\subsubsection{Mismatch Between Benchmarks and Real-World Use}
Among the six real-world capabilities we evaluate, Reviewing Work and Data Structuring are not captured at all by the metrics used in current AI benchmarking. Meanwhile, Summarization and Generation, capabilities widely used by non-technical professionals are only weakly supported. This highlights a consistent mismatch between what AI benchmarks measure and how AI is actually used in practice. The gap suggests that existing evaluation tools under-prioritise the capabilities most relevant to the broader AI user base.

\subsection{Model evaluation}

\begin{figure}[ht]
    \centering
    \includegraphics[width=\textwidth]{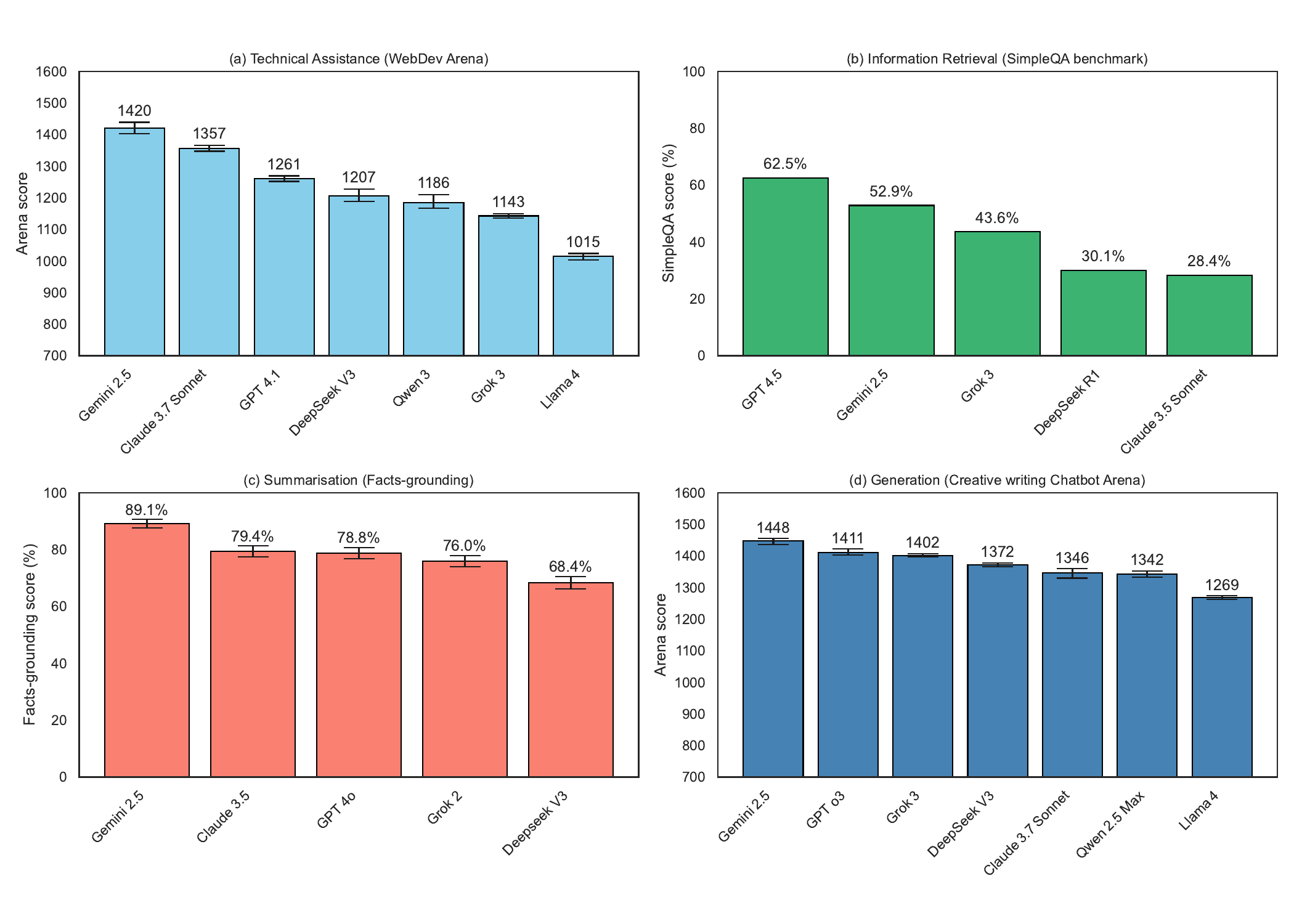}
    \caption{
        Evaluation results for highest ranking models, we only take the highest scoring model from one company, so they may have models that score higher than other companies. Scores are correct as of 12th May 2025. Across four of the six different AI capabilities outlined in this paper. 
       \textbf{(a) Technical Assistance (WebDev Arena):} Elo-style Arena scores, derived from head-to-head matchups in web development tasks, reflect relative performance. Higher values indicate better comparative results: error bars show 95\% confidence intervals. 
        \textbf{(b) Information Retrieval (SimpleQA benchmark):} Accuracy percentage on a benchmark testing factual correctness in short-answer question responses. The y-axis shows the proportion of correct answers; higher values indicate better factual precision. 
        \textbf{(c) Summarization (Facts-grounding benchmark):} Percentage of grounded content in model-generated summaries, based on human judgments of factual consistency with source documents. Error bars indicate 95\% confidence intervals. 
        \textbf{(d) Generation (Creative Writing Arena):} Elo scores from subjective human preferences in open-ended Generation tasks such as storytelling and creative writing. Higher scores denote consistent wins in pairwise comparisons; error bars show 95\% confidence intervals. 
    }

    \label{fig:combined_metric}
\end{figure}

Figure~\ref{fig:combined_metric} presents the performance of the highest-ranking models across four distinct AI capabilities, revealing several notable trends. Most prominently, Gemini 2.5 appears in the top two positions across all four tasks, ranking first in Summarization (89.1\%),  Generation (Elo score of 1458), and Technical Assistance (Elo score of 1420). This consistent high performance highlights Gemini’s versatility and strength across a diverse range of tasks, including factual grounding, creative writing, and Technical Assistance. Anthropic  models also demonstrate strong results: Claude ranks second in Summarization (79.4\%) and second in Technical Assistance (Elo score of 1357).

\section{Discussion}

\subsection{Metrics and Real–World Usage: A Fundamental Mismatch}
The Results section demonstrates how the metrics employed by AI companies align with various functional capabilities. However, most technical reports released by AI companies, along with academic evaluations and industry leaderboards, rely heavily on aggregate benchmark scores to assess model performance \cite{deepseek_release, OpenAI2023, anthropic2025claude37sonnet}. These benchmarks are typically designed to evaluate models on narrowly defined tasks such as multiple-choice exams, coding problems, or symbolic logic puzzles—tasks that are meant to measure traditional conceptions of intelligence as the ability to reason abstractly, solve problems, or retrieve factual knowledge. Despite measuring AI capability in limited tasks, this framing obscures an important reality: LLMs are not just tools for solving bounded tasks with single correct answers—they are often deployed in open-ended, dialogic, and highly contextual workflows. Our analysis shows that, alongside technical tasks like coding and data manipulation, users commonly rely on LLMs for co-authoring emails, providing stylistic feedback, reformatting citations, summarizing long documents, and brainstorming ideas \cite{handa2025economic, humlum2024}. These tasks are better understood not as tests of intelligence, but as forms of collaborative cognition where the AI plays the role of assistant, or editor, where applying the same objective standards to evaluate their strengths and weaknesses proves challenging.

This divergence between benchmark design and everyday use suggests that prevailing evaluations fail to capture the practical, human-centered goals that motivate the adoption of generative AI. By privileging abstract measures of machine intelligence over grounded assessments of utility, these benchmarks risk misrepresenting both the strengths and limitations of current models.This suggests a need to evaluate benchmarks not against abstract tasks alone, but through the lenses of real-world utility—where usefulness, clarity, and time savings take precedence over syntactic correctness or answer matching.

Most public benchmarks focus on technical tasks—code Generation, mathematics, and factual recall—while linguistically rich or socially situated tasks (e.g., tone editing, policy summarization) remain underrepresented. This skew reflects not just a preference for measurable outputs, but a deeper misalignment between benchmark design and the diverse ways people use AI in practice. While these tasks may be easier to score automatically, they capture only a narrow slice of the real-world interactions that users engage in daily—interactions that often involve open-ended reasoning, contextual sensitivity, and human judgment.

This mismatch is particularly evident when considering workforce data. Survey findings indicate that 88\% of employees using AI on the job are non‑technical workers—educators, healthcare staff, and retail associates—compared to just 12\% in software development or engineering roles \cite{DeSmet2024HumanGenerativeAI}. Yet benchmark coverage and public prompt datasets disproportionately reflect the workflows of developers. In Anthropic’s prompt-to-task dataset, for example, nearly two-thirds of high-use prompts relate to Technical Assistance \cite{handa2025economic}. One hypothesis for this imbalance  stems from differences in prompt frequency: developers often engage in multi-turn, iterative interactions with AI tools, whereas non‑technical users may issue fewer but more varied prompts aimed at business communication, summarization, or formatting. The result is a feedback loop in which coding use cases dominate evaluation design, while broader categories of human–AI collaboration remain undervalued, despite their wide-use. However, a study would need to be conducted to confirm this.

\subsection{The Vanishing Human in the Loop}
Although metrics such as HumanEval and MMLU purport to assess model accuracy and reasoning, they depend almost entirely on automated scoring pipelines and, in some cases, on other LLMs for adjudication \cite{miller2024errorbars, he2023blindspots}. This approach contrasts sharply with the human‑mediated evaluation practices that govern professional and academic settings, —such as peer review of scholarly writing, editorial judgment in journalism, or code reviews in software engineering where domain experts judge the quality of summaries, translations, or code reviews \cite{howtodohumaneval}. By eliminating human raters, large‑scale benchmarks can obscure ambiguous or context‑sensitive errors—hallucinations, cultural misinterpretations, or stylistic misalignments—that are readily detected in real‑world use. Reintroducing human judgement into evaluation frameworks, even at the expense of smaller test sets, would more faithfully reflect the collaborative nature of human–AI workflows.

\subsection{Coverage Gaps Across Capabilities}
Our thematic analysis identified six core AI capabilities—Summarization, Technical Assistance, Reviewing Work, Data Structuring, Generation, and Information Retrieval—yet only four of these map onto established benchmarks. Notably, Reviewing Work, which encompasses proofreading, tone adjustment, and structural feedback, appears in 58.9\% of high‑frequency occupational tasks but lacks a dedicated evaluation suite. Similarly, Data Structuring tasks such as citation conversion or table formatting account for 4\% of real‑world prompts yet are unrepresented in mainstream test collections. The absence of metrics for these capabilities not only deprives users of guidance when selecting models but also enables vendors to emphasize strengths in well‑measured areas while neglecting essential collaborative functions.

\subsection{Blind Spots in Current Benchmarks} 
Building on the benchmark reviews in Section~\ref{sec:metrics}, several blind spots emerge that constrain their applicability to real-world workflows. While current benchmark suites offer valuable insights into certain model capabilities, they also exhibit notable blind spots that limit their applicability to real-world workflows. First, the emphasis on Python in Technical Assistance benchmarks—such as HumanEval and MBPP—overrepresents support for a single programming language. In contrast, few benchmarks evaluate assistance with other widely used tools and languages such as R, Excel, LaTeX, or domain-specific scripting environments, despite their prevalence in professional contexts. As a result, models may appear well-rounded in technical performance while providing limited support for non-Python workflows.

Second, most existing benchmarks prioritize correctness while largely overlooking efficiency. For example, a model may return an accurate citation or summary, but the time or cognitive effort required to parse or integrate the output remains unmeasured. In real-world settings, a key value proposition of generative AI is its ability to reduce user workload. Yet benchmarks rarely quantify this benefit, failing to ask: How much faster did the user complete their task? or How many revision cycles were saved?

Finally, there is limited attention paid to interpretability from the user's perspective; an aspect we categorize under subjective criteria. Evaluation pipelines often treat the presence of a correct answer as sufficient, without assessing whether the output is understandable, logically structured, or appropriately formatted for human consumption. This is especially problematic for tasks involving technical documentation, policy analysis, or instructional content, where clarity and usability are as important as correctness. Without incorporating interpretability metrics—such as alignment with domain conventions or ease of follow-up—the utility of model outputs in collaborative settings remains under-evaluated.

Together, these blind spots point to a deeper issue: benchmarks that foreground abstract correctness over contextual relevance risk mischaracterizing what it means for a model to be “useful.” Expanding evaluation frameworks to include tool coverage, task efficiency, and interpretability would offer a more holistic picture of model capability and better align with user needs.

\subsection{Towards Capability–Aligned, Human–Centred Metrics}
To close the gap between benchmark performance and user value, we propose constructing bespoke evaluation suites for each of the six identified capabilities. Each suite should be designed around the five criteria of coherence, accuracy, clarity, relevance, and efficiency, with efficiency defined in terms of time or cognitive effort saved. Tasks must permit multi‑turn interaction, enabling iterative refinement, and should integrate human raters for any dimension that involves subjective judgement. Smaller, carefully curated datasets, transparent annotation protocols, and public prompt repositories will provide more reliable and actionable signals than the sprawling, opaque corpora that currently dominate the field.

This proposal addresses a growing tension in the field: while alignment methods like RLHF have improved how closely model outputs reflect human preferences \cite{bai2022constitutional, ouyang2022training}, evaluation remains largely decoupled from these goals—focusing instead on abstract problem-solving benchmarks that overlook real-world use.

\subsection{Interpreting Leaderboards: Recency and Transparency Effects}

Following the release of DeepSeek-R1 in January 2025, a wave of competing models was announced in rapid succession—OpenAI, Google, Anthropic, Alibaba, and others each launched models in February and March that narrowly outperformed DeepSeek on various benchmark metrics \cite{openai_gpt45,google2025gemini2.5,qwenlm2025qwen2.5max,anthropic2025claude37sonnet}. Each successive release claimed marginal improvements over its predecessors, often by one or two percentage points, suggesting a pattern of incremental gains driven by awareness of previous scores. This phenomenon highlights what we term a "recency advantage," where developers of newer models can fine-tune against published benchmarks and adjust hyperparameters or test-time configurations to outperform the latest leader. However, this advantage does not necessarily reflect genuine architectural advances. 

Compounding this issue is the lack of transparency in evaluation practices. Many performance results—such as those we analyze in this paper—are self-reported by companies without standardized auditing, making it difficult to determine whether models were evaluated under comparable conditions. For example, it is often unclear whether models are run multiple times with only the best-performing result reported, or whether evaluations were specifically optimized to highlight strengths. This leaderboard arms race incentivizes superficial metric gains rather than meaningful improvements aligned with real-world user needs.

Crowd-sourced evaluations such as the Chatbot Arena address some of these issues by incorporating human-in-the-loop comparisons and being run by an external party. However, the platform still lacks full transparency: the prompt set is not publicly disclosed, rater demographics are unknown, and ratings may be influenced by style preferences or familiarity with certain model behaviors. Moreover, the comparative nature of the judgments—asking which model is better rather than how well a model performs against a defined standard—limits interpretability. Recent research has highlighted additional concerns, such as the ability of certain providers to test multiple model variants before public release and selectively disclose performance results, leading to biased Arena scores due to selective disclosure of performance results. Furthermore, proprietary closed models are sampled at higher rates and have fewer models removed from the arena than open-weight and open-source alternatives, resulting in data access asymmetries that can distort leaderboard rankings \cite{singh2025leaderboard}. In the absence of an objective gold standard or fine-grained error analysis, leaderboard rankings risk overstating differences while obscuring common failure modes across models.

A further limitation is that many of these evaluations rely on single-turn prompts. Yet in real-world applications, text-to-text models are rarely used in isolation. Instead, they function within multi-turn, iterative workflows—refining outputs, responding to clarification requests, and adapting to changing context. A single prompt evaluation fails to capture these dynamics, leaving critical aspects of interaction quality unmeasured. To move beyond superficial comparisons, we need benchmark designs that integrate objective reference answers, track error margins, and account for the conversational, evolving nature of human–AI collaboration.

\subsection{Implications, Limitations, and Future Work}
The divergence between benchmark scores and actual user workflows underscores the need to reorient evaluation culture toward human-centred measures of utility. Rather than focusing narrowly on correctness or static tasks, evaluations should consider whether models meaningfully streamline or support the tasks users care about most. Such a shift is essential for informing responsible procurement, deployment, and regulation of generative AI systems—ensuring that advances in model architecture translate into genuine improvements in human productivity and satisfaction.

Although our analysis integrates survey data, real-world usage logs, and benchmark documentation, it is subject to several limitations. First, the process of mapping raw Claude.ai prompts to predefined occupational tasks may oversimplify user intent and overlook nuanced or cross-functional activities that do not fit neatly into O*NET categories. Second, our assessment of benchmark design draws on publicly available documentation; proprietary evaluation protocols and unpublished internal benchmarks may follow different conventions. Third, while we have proposed capability-aligned criteria, we have not yet empirically validated a concrete evaluation suite for any single capability.

Future work should address these limitations by designing, piloting, and refining new benchmarks aligned with the six capabilities and five evaluative criteria proposed in this paper. Each benchmark should adopt a human-in-the-loop methodology. For example, a Reviewing Work benchmark might present professional documents (e.g., emails, reports, or manuscripts) alongside model-generated edits, with expert annotators rating improvements in style, accuracy, and coherence. A Data Structuring benchmark could supply raw reference lists or unformatted tables and ask human judges to assess correctness and efficiency in standard citation formats or tabular layouts. Benchmarks for Summarization, Technical Assistance, Generation, and Information Retrieval should incorporate multi-turn dialogues, domain-specific tasks, and metrics reflecting time saved or cognitive load reduced. By publicly releasing prompts, annotation protocols, and evaluation rubrics, future research can help establish transparent and reproducible standards that better reflect real-world utility and support more human-centered model development.

\section{Conclusion}
This paper has explored how generative AI is being used in practice and identified a meaningful opportunity to better align evaluation metrics with real-world workflows. While current benchmarks provide useful insights into specific model capabilities—particularly in technical domains—they often focus on narrow, isolated tasks that do not capture the full spectrum of how people interact with AI. By analyzing both survey and usage data, we distilled AI use into six key capabilities: Summarization, Technical Assistance, Reviewing Work, Data Structuring, Generation, and Information Retrieval. These capabilities reflect the diverse, collaborative ways users engage with language models across professional and personal contexts.
Rather than proposing a new evaluation framework, our analysis surfaces where existing benchmarks align with these capabilities—and, notably, where they do not. Many capabilities used frequently in real-world tasks, such as Reviewing Work and Data Structuring, currently lack dedicated benchmarks. Even where coverage exists, we find that commonly used metrics often prioritize abstract notions of correctness over qualities that matter to users, such as clarity, efficiency, and contextual relevance. In addition, many benchmarks rely on decontextualized tasks, offering little variation in domain, audience, or user intent. This limits their ability to capture how model performance varies across the rich range of real-world use cases.

To examine these shortcomings, we evaluate current benchmarks through five practical lenses—coherence, accuracy, clarity, relevance, and efficiency—derived from linguistic principles and user expectations. These dimensions are not presented as a new framework, but as tools to reveal blind spots in current practice and identify areas for improvement.

By highlighting these mismatches between benchmark design and user needs, this paper provides a road-map for more human-centered evaluation. This approach supports more meaningful comparisons between models, helps organizations make better-informed decisions, and encourages developers to focus on the capabilities users care about most. As generative AI continues to become embedded in everyday work, improving how we evaluate these systems is essential to ensuring they deliver genuinely helpful experiences.

%% Bibliography

\bibliography{Main_File}

\appendix
\renewcommand{\thetable}{\Alph{section}\arabic{table}}
\setcounter{table}{0}

\section{Appendix: Capability First Analysis}
\begin{longtable}{|p{3.5cm}|p{5cm}|p{4cm}|p{3cm}|}
\caption{Mapping of 21 high-exposure occupational tasks to AI capabilities based on qualitative coding of tasks rated as offering large time savings through ChatGPT assistance. Tasks span 11 occupations and reflect the cognitive processes and outputs best supported by generative AI.}
 \label{tab:ai-capabilities-tasks} \\
\hline
\textbf{Occupation} & \textbf{Job Task} & \textbf{Explanation} & \textbf{Capability} \\
\hline
\endfirsthead

\hline
\textbf{Occupation} & \textbf{Job Task} & \textbf{Explanation} & \textbf{Capability} \\
\hline
\endhead

\hline
\endfoot

\hline
\endlastfoot
        Accountants \& Auditors & Prepare detailed reports on audit findings. & ChatGPT can draft and structure reports based on audit findings. & Generation, Data Structuring \\ \hline
        Accountants \& Auditors & Examine and evaluate financial and information systems, recommending controls to ensure system reliability and data integrity. & ChatGPT can compile reports on financial and IT systems from provided data and text, identifying potential issues and suggesting controls for system reliability and data integrity. & Summarization, Reviewing Work \\ \hline
        Accountants \& Auditors & Prepare, examine, or analyze accounting records, financial statements, or other financial reports to assess accuracy, completeness, and conformance to reporting and procedural standards. & ChatGPT can draft accounting documents and analyze accounting information and financial reports. & Generation \\ \hline
        Accountants \& Auditors & Compute taxes owed and prepare tax returns, ensuring compliance with payment, reporting, or other tax requirements. & ChatGPT can provide guidance on tax legislation, calculate tax liabilities, and generate drafts of tax returns. & Information Retrieval \\ \hline
        Customer Service Rep. & Keep records of customer interactions or transactions, recording details of inquiries, complaints, or comments, as well as actions taken. & ChatGPT can assist with logging and reporting customer contacts based on data from customer support. & Data Structuring \\ \hline
        Customer Service Rep. & Check to ensure that appropriate changes were made to resolve customersâ€™ problems. & ChatGPT can prepare a structured report on whether the actions taken resolved the customer complaint. & Summarization, Generation \\ \hline
        Customer Service Rep. & Contact customers to respond to inquiries or to notify them of claim investigation results or any planned adjustments. & ChatGPT can suggest responses to customer inquiries and complaints. & Generation \\ \hline
        Financial Advisors & Recommend to clients strategies in cash management, insurance coverage, investment planning, or other areas to help them achieve their financial goals. & ChatGPT can develop and formulate financial strategies and plans based on a client's financial situation and goals. & Generation \\ \hline
        Financial Advisors & Implement financial planning recommendations, or refer clients to someone who can assist them with plan implementation. & ChatGPT can provide step-by-step instructions for implementing a client's financial plan and suggest agents who can assist with the implementation. & Generation \\ \hline
        Financial Advisors & Analyze financial information obtained from clients to determine strategies for meeting clientsâ€™ financial objectives. & ChatGPT can suggest and describe suitable financial strategies based on clientsâ€™ financial situations and goals. & Generation \\ \hline
        Financial Advisors & Answer clientsâ€™ questions about the purposes and details of financial plans and strategies. & ChatGPT can suggest answers to typical questions about financial plans and strategies. & Information Retrieval \\ \hline
        HR Professionals & Inform job applicants of details such as duties and responsibilities, compensation, benefits, schedules, working conditions, or promotion opportunities. & ChatGPT can generate detailed descriptions of jobs and employment terms. & Summarization \\ \hline
        HR Professionals & Interpret and explain human resources policies, procedures, laws, standards, or regulations. & ChatGPT can interpret and explain complex HR policies and regulations in easily accessible language. & Summarization \\ \hline
        IT Support & Answer user inquiries regarding computer software or hardware operation to resolve problems. & ChatGPT can provide step-by-step instructions for solving typical hardware and software problems. & Technical Assistance \\ \hline
        IT Support & Read technical manuals, confer with users, or conduct computer diagnostics to investigate and resolve problems or to provide technical assistance and support. & ChatGPT can summarize technical manuals and assist with technical support by suggesting questions to users and possible solutions. & Summarization, Technical Assistance \\ \hline
        IT Support & Maintain records of daily data communication transactions, problems and remedial actions taken, or installation activities. & ChatGPT can structure log files and notes into coherent reports. & Summarization \\ \hline
        Journalists & Write commentaries, columns, or scripts. & ChatGPT can generate drafts, suggest changes, and provide ideas for articles, etc. & Generation, Reviewing Work \\ \hline
        Journalists & Examine news items of local, national, and international significance to determine topics to address, or obtain assignments from editorial staff members. & ChatGPT can analyze and summarize news content and suggest topics to cover. & Summarization \\ \hline
        Journalists & Analyze and interpret news and information received from various sources to broadcast the information. & ChatGPT can analyze, summarize, and translate news from various sources. & Summarization, Information Retrieval \\ \hline
        Legal Professionals & Prepare affidavits or other documents, such as legal correspondence, and organize and maintain documents in paper or electronic filing system. & ChatGPT can suggest templates and drafts for legal documents and provide guidance on filing. & Generation, Information Retrieval \\ \hline
        Legal Professionals & Prepare legal documents, including briefs, pleadings, appeals, wills, contracts, and real estate closing statements. & ChatGPT can deliver drafts of legal documents based on entered details. & Generation \\ \hline
        Marketing Professionals & Prepare reports of findings, illustrating data graphically and translating complex findings into written text. & ChatGPT can write and structure reports from data and text, and can also suggest presentation forms for data. & Generation \\ \hline
        Office Clerks & Communicate with customers, employees, and other individuals to answer questions, disseminate or explain information, take orders, and address complaints. & ChatGPT can suggest responses to typical inquiries, complaints, and orders. & Generation \\ \hline
        Office Clerks & Compile, copy, sort, and file records of office activities, business transactions, and other activities. & ChatGPT can prepare records following complex instructions and assist with filing and sorting documents by summarizing and editing text. & Reviewing Work, Summarization \\ \hline
        Office Clerks & Compute, record, and proofread data and other information, such as records or reports. & ChatGPT can prepare and check records and reports based on predefined guidelines. & Generation \\ \hline
        Software Developers & Write, analyze, review, and rewrite programs, using workflow chart and diagram, and applying knowledge of computer capabilities, subject matter, and symbolic logic. & ChatGPT can assist with writing code and analyzing errors in programs based on software developers' preferences and program outputs. & Technical Assistance \\ \hline
        Software Developers & Correct errors by making appropriate changes and rechecking the program to ensure that the desired results are produced. & ChatGPT can identify code errors and suggest corrections and checks based on error messages and other program outputs. & Technical Assistance \\ \hline
        Software Developers & Perform or direct revision, repair, or expansion of existing programs to increase operating efficiency or adapt to new requirements. & ChatGPT can provide code suggestions for auditing, debugging, and extending programs, and can also suggest ways to optimize the code. & Technical Assistance, Information Retrieval \\ \hline
        Software Developers & Conduct trial runs of programs and software applications to be sure they will produce the desired information and that the instructions are correct. & ChatGPT can suggest code changes and debug programs, as well as explain program output in a reader-friendly format. & Technical Assistance, Summarization \\ \hline
        Software Developers & Consult with and assist computer operators or system analysts to define and resolve problems in running computer programs. & ChatGPT can identify code errors and suggest corrections based on error messages, program output, and user input. & Technical Assistance \\ \hline
        Teachers & Adapt teaching methods and instructional materials to meet studentsâ€™ varying needs and interests. & ChatGPT can tailor teaching methods and materials based on each student's learning style and interests. & Generation \\ \hline
        Teachers & Prepare objectives and outlines for courses of study, following curriculum guidelines or requirements of states and schools. & ChatGPT can suggest and structure learning objectives and courses in accordance with curricula or similar requirements. & Generation \\ \hline
\end{longtable}

\section{Appendix: Metric Evaluation}
\begin{longtable}{|p{1.5cm}|p{1.5cm}|p{2.7cm}|p{2.7cm}|p{2.7cm}|p{2.7cm}|p{1.7cm}|}
\caption{Comparison of AI Benchmarks by  Capability and  5 Evaluation Criteria} \label{tab:appendix-metric-eval} \\
\hline

\textbf{Metric} & \textbf{Capability} & \textbf{Coherence} & \textbf{Accuracy} & \textbf{Clarity} & \textbf{Relevance} & \textbf{Efficiency} \\
\hline
\endfirsthead
\hline
\textbf{Metric} & \textbf{Capability} & \textbf{Coherence} & \textbf{Accuracy} & \textbf{Clarity} & \textbf{Relevance} & \textbf{Efficiency} \\
\hline
\endhead

MMLU & Information Retrieval & Questions follow a multiple-choice format but lack natural dialogue or conversational flow & 6.5\% of questions contain incorrect answers & Format is rigid; only exact answer options are considered correct & Focused on subjects common in the US education system & Not explicitly measured \\ \hline

AIME & Information Retrieval & Questions ask for one-off numerical answers, with little explanation or reasoning & Based on official math competition questions & Answers are expected to be exact numbers & Covers only mathematics problems & Measured indirectly through response time \\ \hline

Codeforces & Technical Assistance & Tasks are described in plain English programming problems & Performance drops for questions written after the model’s training period & Output is executable code that can be used by a developer & All tasks are code-related, most suited for C++ and Python & Time to solve is measured \\ \hline

GPQA & Information Retrieval & Standard multiple-choice questions without much context & 36\% of questions have disagreement over the correct answer & Exact formatting of answers is required & Covers only biology, chemistry, and physics & Not explicitly measured \\ \hline

SWE-bench & Technical Assistance & Based on real bug reports from software projects & 32\% of solutions used information found in public comments; 31\% had questionable test cases & Requires the model to suggest a working patch that passes automated tests & Focuses entirely on Python software development & Not explicitly measured \\ \hline

FACTS Grounding & Summari\-zation & Large text passages are followed by detailed factual questions & Model answers are rated by a different model or human reviewer, with 62–85\% agreement & Answers must be understandable by other models or humans & Based on dense material such as law, medicine, or STEM writing & Not explicitly measured \\ \hline

MRCR & Summari\-zation & Tasks include similar sentences where models must retrieve the correct one & Measures whether the correct sentence is chosen from among distractors & Emphasizes accurate retrieval over natural phrasing & Based on synthetic data but intended to reflect real-world understanding & Not explicitly measured \\ \hline

HumanEval & Technical Assistance & Each task includes a function description in plain English & Model is judged on whether its function passes test cases & Code does not need to be clear, only correct & All tasks are in Python & Not explicitly measured \\ \hline

SimpleQA & Information Retrieval & Short factual questions that expect short factual answers & 94.4\% agreement between original reviewers and a third reviewer & Prioritizes clear, correct, concise answers & Covers diverse topics from science to culture & Not explicitly measured \\ \hline

GSM8K & Information Retrieval & Math problems are written in plain language and require multi-step reasoning & About 50\% of questions were incorrect in early versions; improved in later release & Focus is on producing the correct final answer, not explanation quality & Covers elementary and middle school mathematics & Not explicitly measured \\ \hline

Math-500 & Information Retrieval & Involves complex math questions requiring multi-step logic & No known major errors in question set & Explanation quality is not rated & Focused entirely on mathematics & Not explicitly measured \\ \hline

Bird-SQL & Technical Assistance & Natural language questions must be converted into SQL code & Some inconsistencies in dataset, but no published error rate & No penalties for unclear output & Covers a wide range of database queries & Measures efficiency using a custom score \\ \hline

LiveCode\-Bench & Technical Assistance & Coding tasks from popular programming websites & Only uses tasks published after model training period & Focus is on whether code runs correctly, not how it’s written & All problems are Python-based & Not explicitly measured \\ \hline

Humanity’s Last Exam & Information Retrieval & Combination of multiple-choice and short-answer questions & Written by experts, including PhDs & Short answers must match exactly to count as correct & Wide coverage, including niche historical and technical content & Not explicitly measured \\ \hline

MMLU-Pro & Information Retrieval & Similar to MMLU but with more answer options per question & Reviewed for robustness with revised questions & Same rigid answer format as MMLU & Broader focus beyond US education & Not explicitly measured \\ \hline

AGIEval & Information Retrieval & Objective questions only; no open-ended responses & Based on official entrance exams like SAT and LSAT & Only exact matches are scored as correct & Covers diverse academic entrance exams & Not explicitly measured \\ \hline

CRUXEval & Technical Assistance & Models must predict input or output of a Python function & Functions adapted from tutorials & Clarity not judged; correctness is all that matters & Functions are short and educational & Not explicitly measured \\ \hline

Writing\-Bench & Generation & Models generate full passages in various writing styles & Judged by a trained model validated against human scores & Penalizes vague or incoherent writing & Includes creative, persuasive, technical, and informative writing & Not explicitly measured \\ \hline

WebDev Arena & Technical Assistance & Prompts ask models to build entire websites or interfaces & No gold standard; judged by human preferences & Clarity depends on how useful output is to humans & Tasks involve real-world web development problems & Efficiency inferred from human preferences \\ \hline

Chatbot Arena & Generation & Creative prompts (e.g., poems, stories) are given to models & Responses compared by direct human voting & Writing quality judged indirectly via preference & Focused on imaginative, creative writing & Not explicitly measured \\ \hline
\end{longtable}

\section{Appendix: Prompts for all applicable capabilities}
\setlength\LTleft{0pt} 
\setlength\LTright{0pt}
\renewcommand\arraystretch{1.2}

\begin{longtable}{|p{3.5cm}|p{2.6cm}|p{1.1cm}|p{0.8cm}|p{3cm}|p{5cm}|}
\caption{Top 100 tasks given by users or Anthropic, showing the aligned AI capability, example prompts, and the percentage of prompts associated with each task. Tasks are sorted by frequency, with the highest-ranked tasks appearing first.} \\ 
\hline
\label{tab:top100-anthropic-tasks} 

\textbf{Task} & \textbf{Profession} & \textbf{Pct of prompts} & \textbf{Rank} & \textbf{AI Capability} & \textbf{Example Prompts} \\
\hline
\endfirsthead
\hline
\textbf{Task} & \textbf{Profession} & \textbf{Pct of prompts} & \textbf{Rank} & \textbf{AI Capability} & \textbf{Example Prompts} \\
\hline
\endhead
Modify existing software to correct errors, to adapt it to new hardware, or to upgrade interfaces and improve performance. & Software developers, systems software & 4.79 & 1 & Technical Assistance, Reviewing Work & \begin{itemize}
  \item My Python program runs really slowly when i open a large file. How can i optimise it?
  \item Can you check this c++ code for compatibility with newer hardware?
\end{itemize} \\ \hline
Correct errors by making appropriate changes and rechecking the program to ensure that the desired results are produced. & Computer programmers & 3.0 & 2 & Reviewing Work, Technical Assistance & \begin{itemize}
  \item Can you help me debug this for loop? It's not producing the correct output.
  \item I'm getting a 'null pointer exception' in Java. How do i fix it?
\end{itemize} \\ \hline
Modify existing software to correct errors, allow it to adapt to new hardware, or to improve its performance. & Software developers, applications & 2.52 & 3 & Reviewing Work, Technical Assistance & \begin{itemize}
  \item Identify performance bottlenecks in my TensorFlow code.
  \item Update my app to run efficiently on M1 Macs
\end{itemize} \\ \hline
Perform initial debugging procedures by reviewing configuration files, logs, or code pieces to determine breakdown source. & Software quality assurance engineers and testers & 2.01 & 4 & Technical Assistance, Reviewing Work & \begin{itemize}
  \item Can you find common causes for 'segfault' in c programs?
  \item Here's my server log — what caused the crash?
\end{itemize} \\ \hline
Perform routine system administrative functions such as troubleshooting, back-ups, and upgrades. & Bioinformatics technicians & 1.81 & 5 & Technical Assistance & \begin{itemize}
  \item How do I automate backups of my Linux server every night?
\end{itemize} \\ \hline
Diagnose, troubleshoot, and resolve hardware, software, or other network and system problems, and replace defective components when necessary. & Network and computer systems administrators & 1.56 & 6 & Technical Assistance & \begin{itemize}
  \item My wi-fi disconnects randomly. What are some possible causes?
\end{itemize} \\ \hline
Write new programs or modify existing programs to meet customer requirements, using current programming languages and technologies. & Data warehousing specialists & 1.46 & 7 & Technical Assistance & \begin{itemize}
  \item Write SQL code to create a sales summary table by region.
\end{itemize} \\ \hline
Select and edit documents for publication and display, applying knowledge of subject, literary expression, and presentation techniques. & Archivists & 1.41 & 8 & Reviewing Work, Generation, Summarization & \begin{itemize}
  \item Can you edit this paragraph to make it more formal and clear?
  \item Write an exhibition description for this archived letter
  \item Summarise this historical document in 100 words.
\end{itemize} \\ \hline
Review and analyze computer printouts and performance indicators to locate code problems, and correct errors by correcting codes. & Computer systems analysts & 1.35 & 9 & Reviewing Work, Technical Assistance & \begin{itemize}
  \item Here's the output from my script. Why is it failing?
  \item My server report shows high cpu usage at midnight. What could be the cause?
\end{itemize} \\ \hline
Write, analyze, review, and rewrite programs, using workflow chart and diagram, and applying knowledge of computer capabilities, subject matter, and symbolic logic. & Computer programmers & 1.24 & 10 & Reviewing Work, Technical Assistance & \begin{itemize}
  \item Review this flowchart and suggest improvements to the algorithm.
  \item Generate a Python program that implements a binary search tree.
\end{itemize} \\ \hline
Design, build, or maintain web sites, using authoring or scripting languages, content creation tools, management tools, and digital media. & Web developers & 1.19 & 11 & Technical Assistance & \begin{itemize}
  \item Help me write javascript to make this image carousel auto-scroll.
\end{itemize} \\ \hline
Write, update, and maintain computer programs or software packages to handle specific jobs such as tracking inventory, storing or retrieving data, or controlling other equipment. & Computer programmers & 1.15 & 12 & Technical Assistance& \begin{itemize}
  \item Write a Python script that tracks inventory in a csv file.
\end{itemize} \\ \hline
Prepare, rewrite and edit copy to improve readability, or supervise others who do this work. & Editors & 1.0 & 13 & Reviewing Work, Generation & \begin{itemize}
  \item Edit this paragraph to be more concise and academic.
  \item Write an article to appeal to high school readers.
\end{itemize} \\ \hline
Determine sources of web page or server problems, and take action to correct such problems. & Web administrators & 0.89 & 14 & Reviewing Work, Technical Assistance, Information Retrieval & \begin{itemize}
  \item Look at these logs, can you tell what's wrong with my apache config?
  \item My website is showing a 500 error. What could be causing it?
  \item What are common causes of slow server load times?
\end{itemize} \\ \hline
Edit, standardize, or make changes to material prepared by other writers or establishment personnel. & Technical writers & 0.85 & 15 & Reviewing Work, Summarization & \begin{itemize}
  \item Edit this user manual section to follow our company style guide.
  \item Summarise the key points of this technical spec document.
\end{itemize} \\ \hline
Review class material with students by discussing text, working solutions to problems, or Reviewing Worksheets or other assignments. & Tutors & 0.79 & 16 & Generation, Summarization & \begin{itemize}
  \item Create a worksheet that will test year 7 students english ability
  \item Summarise this chapter on plant biology for year 10 students.
\end{itemize} \\ \hline
Modify existing programs to enhance efficiency. & Computer numerically controlled machine tool programmers, metal and plastic & 0.78 & 17 & Technical Assistance & \begin{itemize}
  \item Optimise this g-code script for faster cnc milling
\end{itemize} \\ \hline
Edit or rewrite existing copy as necessary, and submit copy for approval by supervisor. & Copy writers & 0.76 & 18 & Reviewing Work & \begin{itemize}
  \item Check this sales pitch paragraph for tone and clarity.
\end{itemize} \\ \hline
Develop or apply data mining and machine learning algorithms. & Bioinformatics technicians & 0.63 & 19 & Technical Assistance & \begin{itemize}
  \item Help me write a code for k-means clustering.
\end{itemize} \\ \hline
Write supporting code for web applications or web sites. & Web developers & 0.6 & 20 & Technical Assistance & \begin{itemize}
  \item Write javascript code to validate a contact form
\end{itemize} \\ \hline
Organize material and complete writing assignment according to set standards regarding order, clarity, conciseness, style, and terminology. & Technical writers & 0.59 & 21 & Generation, Reviewing Work & \begin{itemize}
  \item Write a concise summary for the introduction of this user manual.
  \item Can you revise this technical paragraph to improve clarity?
\end{itemize} \\ \hline
Modify existing databases and database management systems or direct programmers and analysts to make changes. & Database administrators & 0.55 & 22 & Technical Assistance & \begin{itemize}
  \item How do i update all customer emails in a PostgreSQL database?
\end{itemize} \\ \hline
Provide private instruction to individual or small groups of students to improve academic performance, improve occupational skills, or prepare for academic or occupational tests. & Tutors & 0.54 & 23 & Reviewing Work, Generation, Summarization & \begin{itemize}
  \item Check my short essay on climate change for grammar and structure.
  \item Write a lesson plan for a student with the following needs
  \item Summarise the key points of this shakespeare sonnet.
\end{itemize} \\ \hline
Guide clients in the development of skills or strategies for dealing with their problems. & Mental health counselors & 0.53 & 24 & Generation& \begin{itemize}
  \item Can you generate daily affirmations for anxiety?
\end{itemize} \\ \hline
Write original or adapted material for dramas, comedies, puppet shows, narration, or other performances. & Actors & 0.5 & 25 & Generation & \begin{itemize}
  \item Write a comedic monologue about being stuck in traffic.
\end{itemize} \\ \hline
Write advertising copy for use by publication, broadcast, or internet media to promote the sale of goods and services. & Copy writers & 0.49 & 26 & Generation, Reviewing Work & \begin{itemize}
  \item Create a product description for a smart water bottle
  \item Can you rewrite this ad copy to make it more persuasive?
\end{itemize} \\ \hline
None &  & 0.48 & 27 &  &  \\ \hline
Read from sacred texts such as the bible, torah, or koran. & Clergy & 0.46 & 28 & Summarization, Information Retrieval & \begin{itemize}
  \item Summarise the parable of the good samaritan.
  \item What is the historical background of the torah?
\end{itemize} \\ \hline
Analyze information to determine, recommend, and plan layout, including type of computers and peripheral equipment modifications. & Computer hardware engineers & 0.43 & 29 & Technical Assistance & \begin{itemize}
  \item How do i plan a workstation layout for high-throughput computing?
\end{itemize} \\ \hline
Compute dimensions, areas, volumes, and weights. & Patternmakers, wood & 0.42 & 30 & Information Retrieval & \begin{itemize}
  \item Calculate the volume of a cylinder with radius 10cm and height 25cm.
\end{itemize} \\ \hline
Conduct logical analyses of business, scientific, engineering, and other technical problems, formulating mathematical models of problems for solution by computers. & Computer and information research scientists & 0.41 & 31 & Technical Assistance, Information Retrieval & \begin{itemize}
  \item Help me write a simulation model for disease spread using Python.
  \item Please explain the travelling salesman problem and how it relates to graph theory
\end{itemize} \\ \hline
Prepare and deliver lectures to undergraduate or graduate students on topics such as how to speak and write a foreign language and the cultural aspects of areas where a particular language is used. & Foreign language and literature teachers, postsecondary & 0.38 & 32 & Information Retrieval, Generation & \begin{itemize}
  \item What are common words i should know before visiting korea?
  \item Create a lesson plan about chinese idioms.
\end{itemize} \\ \hline
Develop factors such as themes, plots, characterizations, psychological analyses, historical environments, action, and dialogue, to create material. & Poets, lyricists and creative writers & 0.37 & 33 & Generation, Reviewing Work, Summarization & \begin{itemize}
  \item Write a poem about lost time using abab rhyme
  \item Review this character sketch and suggest improvements.
  \item Summarise the main themes of hamlet
\end{itemize} \\ \hline
Analyze system performance or operational requirements. & Photonics engineers & 0.36 & 34 & Reviewing Work, Technical Assistance & \begin{itemize}
  \item Evaluate this optical system for energy loss.
  \item How can i improve the performance of a photonic integrated circuit?
\end{itemize} \\ \hline
Read written materials, such as legal documents, scientific works, or news reports, and rewrite material into specified languages. & Interpreters and translators & 0.35 & 35 &  Information Retrieval, Generation & \begin{itemize}
  \item Translate this paragraph into spanish
  \item What's the difference between 'rapport' and 'relationship' in spanish?
\end{itemize} \\ \hline
Analyze organizational, occupational, and industrial data to facilitate organizational functions and provide technical information to business, industry, and government. & Compensation, benefits, and job analysis specialists & 0.35 & 36 & Data Structuring, Summarization & \begin{itemize}
  \item Convert these notes on who reports to who, into an org chart
  \item Summarise the key compensation trends in the tech industry.
\end{itemize} \\ \hline
Review software documentation to ensure technical accuracy, compliance, or completeness, or to mitigate risks. & Software quality assurance engineers and testers & 0.34 & 37 & Technical Assistance & \begin{itemize}
  \item Check this API documentation for clarity and completeness.
\end{itemize} \\ \hline
Break systems into their component parts, assign numerical values to each component, and examine the mathematical relationships between them. & Operations research analysts & 0.33 & 38 & Information Retrieval, Technical Assistance & \begin{itemize}
  \item What are standard methods for queuing theory analysis?
  \item Model this supply chain problem using linear programming.
\end{itemize} \\ \hline
Write, design, or edit web page content, or direct others producing content. & Web developers & 0.32 & 39 & Generation, Reviewing Work & \begin{itemize}
  \item Create content for an about us page for a small bakery.
  \item Check this blog draft for grammar and tone.
\end{itemize} \\ \hline
Compile and write documentation of program development and subsequent revisions, inserting comments in the coded instructions so others can understand the program. & Computer programmers & 0.31 & 40 & Technical Assistance, Reviewing Work & \begin{itemize}
  \item Write docstrings for this Python function.
  \item Check if these comments clearly explain what the code does.
\end{itemize} \\ \hline
Use models to simulate the behavior of animated objects in the finished sequence. & Multimedia artists and animators & 0.31 & 41 & Technical Assistance & \begin{itemize}
  \item Write code to simulate realistic movement for a cloth blowing in the wind.
\end{itemize} \\ \hline
Observe and evaluate students' work to determine progress, provide feedback, and make suggestions for improvement. & Vocational education teachers, postsecondary & 0.31 & 42 & Reviewing Work, Summarization & \begin{itemize}
  \item Evaluate this student's woodworking project based on provided rubric.
  \item Summarise overall student progress from these weekly reports.
\end{itemize} \\ \hline
Analyze user needs and software requirements to determine feasibility of design within time and cost constraints. & Software developers, applications & 0.3 & 43 & Technical Assistance, Information Retrieval & \begin{itemize}
  \item How do i assess whether a feature request can be implemented in under 2 weeks?
  \item What features do users prioritise in this type of software.
\end{itemize} \\ \hline
Determine supplementary virtual features, such as currency, item catalog, menu design, and audio direction. & Video game designers & 0.3 & 44 & Generation, Information Retrieval & \begin{itemize}
  \item Design a virtual currency system that discourages hoarding.
  \item What are ways successful video game franchises have organised their menu layout to be less annoying for the players?
\end{itemize} \\ \hline
Store, retrieve, and manipulate data for analysis of system capabilities and requirements. & Software developers, applications & 0.29 & 45 & Technical Assistance & \begin{itemize}
  \item Write Python code to load and clean JSON data
\end{itemize} \\ \hline
Interact with clients to assist them in gaining insight, defining goals, and planning action to achieve effective personal, social, educational, and vocational development and adjustment. & Clinical psychologists & 0.29 & 46 & Generation & \begin{itemize}
  \item Suggest 3 goals for a cbt therapy plan for anxiety.
\end{itemize} \\ \hline
Test machines, components, materials, or products to determine characteristics such as performance, strength, or response to stress. & Mechanical engineering technologists & 0.28 & 47 & Technical Assistance, Information Retrieval & \begin{itemize}
  \item Generate a script to log temperature and pressure test data.
  \item How do i measure tensile strength in composite materials?
\end{itemize} \\ \hline
Compile reports, charts, or graphs that describe and interpret findings of analyses. & Statistical assistants & 0.27 & 48 & Summarization, Generation, Technical Assistance & \begin{itemize}
  \item Summarise this dataset's key trends in 3 bullet points.
  \item Generate a short paragraph for the executive summary section of this report.
  \item Create a graph of age distribution from this csv
\end{itemize} \\ \hline
Compose letters in reply to correspondence concerning such items as requests for merchandise, damage claims, credit information requests, delinquent accounts, incorrect billing, or unsatisfactory service. & Correspondence clerks & 0.26 & 49 & Reviewing Work, Generation & \begin{itemize}
  \item Check this response letter for tone and grammar.
  \item Write a polite response to a customer requesting a refund.
\end{itemize} \\ \hline
Design databases to support business applications, ensuring system scalability, security, performance and reliability. & Database architects & 0.25 & 50 & Information Retrieval, Technical Assistance  & \begin{itemize}
\item What best practices should I follow to ensure high performance and reliability in a cloud-based SQL database used by a financial application?
\item Can you help me model a normalized relational database for a hospital management system
\end{itemize} \\ \hline
Confer with clients regarding the nature of the information processing or computation needs a computer program is to address. & Computer systems analysts & 0.25 & 51 & Technical Assistance & \begin{itemize}
  \item Help me evaluate client requirements and translate them into specs.
\end{itemize} \\ \hline
Reformat documents, moving paragraphs or columns. & Word processors and typists & 0.25 & 52 & Data Structuring & \begin{itemize}
  \item Move all paragraphs starting with 'summary' to the end of the document
\end{itemize} \\ \hline
Consult with multiple stakeholders to define requirements and implement online features. & Video game designers & 0.24 & 53 & Generation, Technical Assistance & \begin{itemize}
  \item Suggest online features for a multiplayer puzzle game.
  \item How do I implement a player ranking system in unity?
\end{itemize} \\ \hline
Test programs or databases, correct errors, and make necessary modifications. & Database administrators & 0.24 & 54 & Reviewing Work, Technical Assistance & \begin{itemize}
  \item Review this database update script for problems.
  \item Write code that will help me run tests on my database to determine any vulnerabilities
\end{itemize} \\ \hline
Provide technical guidance or support for the development or troubleshooting of systems. & Computer systems engineers/architects & 0.24 & 55 & Technical Assistance, Information Retrieval & \begin{itemize}
  \item Help me design an API server that won't crash under load?
  \item What are common causes of memory leaks in node.js?
\end{itemize} \\ \hline
Analyze operations to evaluate performance of a company or its staff in meeting objectives or to determine areas of potential cost reduction, program improvement, or policy change. & Chief executives & 0.24 & 56 & Summarization, Reviewing Work & \begin{itemize}
  \item Summarize the key weaknesses in our quarterly performance report.
  \item Evaluate this staff productivity data and suggest improvements.
\end{itemize} \\ \hline
Test, maintain, and monitor computer programs and systems, including coordinating the installation of computer programs and systems. & Computer systems analysts & 0.24 & 57 & Technical Assistance & \begin{itemize}
  \item How do I monitor server health in real-time?
\end{itemize} \\ \hline
Explain general financial topics to clients, such as credit report ratings, bankruptcy laws, consumer protection laws, wage attachments, or collection actions. & Credit counselors & 0.23 & 58 & Information Retrieval & \begin{itemize}
  \item What does a credit score of 620 mean?
\end{itemize} \\ \hline
Instruct individuals in career development techniques such as job search and application strategies, resume writing, and interview skills. & Educational, guidance, school, and vocational counselors & 0.23 & 59 & Generation, Reviewing Work & \begin{itemize}
  \item Generate 5 interview questions for a software developer role.
  \item Review this resume and suggest changes
\end{itemize} \\ \hline
Analyze organic or inorganic compounds to determine chemical or physical properties, composition, structure, relationships, or reactions, using chromatography, spectroscopy, or spectrophotometry techniques. & Chemists & 0.22 & 60 & Information Retrieval, Summarization & \begin{itemize}
  \item What is the procedure for running a uv-vis spectroscopy test?
  \item Summarise the findings of this report
\end{itemize} \\ \hline
Plan, evaluate, and revise curricula, course content, course materials, and methods of instruction. & Area, ethnic, and cultural studies teachers, postsecondary & 0.22 & 61 & Summarization, Generation, Information Retrieval, Reviewing Work & \begin{itemize}
  \item Summarise the objectives of this course in 3 bullet points.
  \item Generate lesson plans for a 6-week course on indigenous literatures.
  \item Explain the effects of colonisation on different cultures.
  \item Evaluate this syllabus for coverage of key asian american history topics.
\end{itemize} \\ \hline
Consult with and assist computer operators or system analysts to define and resolve problems in running computer programs. & Computer programmers & 0.22 & 62 & Technical Assistance, Reviewing Work & \begin{itemize}
  \item Generate Python code that is able to handle memory leaks.
  \item Read through my code and identify any issues with it that may cause issues
\end{itemize} \\ \hline
Prepare or edit organizational publications, such as employee newsletters or stockholders' reports, for internal or external audiences. & Public relations specialists & 0.21 & 63 & Generation, Reviewing Work, Summarization, Data Structuring & \begin{itemize}
  \item Write a newsletter update on our recent merger
  \item Review this shareholder report draft for clarity.
  \item Summarise key takeaways from our annual report for employees.
\end{itemize} \\ \hline
Develop, implement, modify, and document recordkeeping and accounting systems, making use of current computer technology. & Accountants & 0.21 & 64 & Data Structuring, Technical Assistance & \begin{itemize}
  \item Format these notes into a table that i can paste into excel.
  \item How do i implement gaap-compliant reports in quickbooks?”
\end{itemize} \\ \hline
Generate standard or custom reports summarizing business, financial, or economic data for review by executives, managers, clients, and other stakeholders. & Business intelligence analysts & 0.2 & 65 & Summarization, Data Structuring& \begin{itemize}
  \item Summarise the findings from this financial paper
  \item Take the data from this spreadsheet and put it into paragraph form
\end{itemize} \\ \hline
Participate in the work of subordinates to facilitate productivity or to overcome difficult aspects of work. & First-line supervisors of office and administrative support workers & 0.2 & 66 & Reviewing Work, Technical Assistance, Generation & \begin{itemize}
  \item Review this task workflow and suggest improvements for speed.
  \item What are some ways to automate excel data entry tasks?
  \item What are some ways to imrpove office productivity?
\end{itemize} \\ \hline
Interpret and analyze policies, public issues, legislation, or the operations of governments, businesses, and organizations. & Political scientists & 0.2 & 67 & Summarization, Reviewing Work, Information Retrieval & \begin{itemize}
  \item Summarise the main arguments in this policy proposal
  \item Review this op-ed on healthcare reform for bias.
  \item What does the eu ai act regulate?
\end{itemize} \\ \hline
Monitor database performance and perform any necessary maintenance, upgrades, or repairs. & Bioinformatics technicians & 0.2 & 68 & Technical Assistance & \begin{itemize}
  \item How do i write a PostgreSQL that will give me the data with the quickest downtime?
\end{itemize} \\ \hline
Address the relationships of quantities, magnitudes, and forms through the use of numbers and symbols. & Mathematicians & 0.2 & 69 & Information Retrieval & \begin{itemize}
  \item What's the difference between eigenvalues and singular values?
\end{itemize} \\ \hline
Design, develop and modify software systems, using scientific analysis and mathematical models to predict and measure outcome and consequences of design. & Software developers, applications & 0.19 & 70 & Technical Assistance & \begin{itemize}
  \item Model a queueing system in Python using discrete event simulation.
\end{itemize} \\ \hline
Write articles, manuals, and other publications, and assist in the distribution of promotional literature about facilities and programs. & Education administrators, elementary and secondary school & 0.19 & 71 & Generation & \begin{itemize}
  \item Write a brochure paragraph about our new after school care program
\end{itemize} \\ \hline
Review instructional content, methods, and student evaluations to assess strengths and weaknesses, and to develop recommendations for course revision, development, or elimination. & Adult basic and secondary education and literacy teachers and instructors & 0.19 & 72 & Reviewing Work, Summarization & \begin{itemize}
  \item Evaluate this adult literacy curriculum for engagement.
  \item Summarise the curriculum for this adult education class
\end{itemize} \\ \hline
Select material most pertinent to presentation, and organize this material into appropriate formats. & Broadcast news analysts & 0.19 & 73 & Summarization & \begin{itemize}
  \item Summarise this press release into a 2-minute news script
\end{itemize} \\ \hline
Collaborate with system architects, software architects, design analysts, and others to understand business or industry requirements. & Database architects & 0.19 & 74 & Information Retrieval, Technical Assistance & \begin{itemize}
  \item What data model suits a large-scale medical records system?
  \item Help me define schema constraints for multi-user access.
\end{itemize} \\ \hline
Review published materials and recommend revisions or changes in scope, format, content, and methods of reproduction and binding. & Technical writers & 0.18 & 75 & Reviewing Work, Generation & \begin{itemize}
  \item Check this user guide for outdated references.
  \item Write installation steps for the new software version
\end{itemize} \\ \hline
Design database applications, such as interfaces, data transfer mechanisms, global temporary tables, data partitions, and function-based indexes to enable efficient access of the generic database structure. & Database architects & 0.18 & 76 & Technical Assistance & \begin{itemize}
  \item Write code that uses function-based indexing to improve query speed.
\end{itemize} \\ \hline
Conduct materials test and analysis using tools and equipment and applying engineering knowledge. & Civil engineering technicians & 0.18 & 77 & Information Retrieval & \begin{itemize}
  \item What astm standards apply to asphalt testing?
\end{itemize} \\ \hline
Use mediation techniques to facilitate communication between disputants, to further parties' understanding of different perspectives, and to guide parties toward mutual agreement. & Arbitrators, mediators, and conciliators & 0.18 & 78 & Information Retrieval, Generation, Reviewing Work & \begin{itemize}
  \item What are some evidence backed ways to ensure a successful mediation
  \item Generate 3 open-ended questions to encourage dialogue between parties.
  \item Evaluate this mediation script for balance and neutrality
\end{itemize} \\ \hline
Organize information for publication and for other means of dissemination, such as use in cd-roms or internet sites. & Historians & 0.18 & 79 & Summarization, Data Structuring & \begin{itemize}
  \item Summarise this archive record
  \item Organise this list of events in the cold war into a chronological order.
\end{itemize} \\ \hline
Prepare and deliver lectures to undergraduate or graduate students on topics such as anatomy, therapeutic recreation, and conditioning theory. & Recreation and fitness studies teachers, postsecondary & 0.17 & 80 & Information Retrieval, Generation & \begin{itemize}
  \item What are the key physiological effects of interval training.
  \item Write lecture notes on muscle anatomy for undergraduate students.
\end{itemize} \\ \hline
Plan parties or other special events and services. & Hosts and hostesses, restaurant, lounge, and coffee shop & 0.17 & 81 & Generation & \begin{itemize}
  \item Generate a birthday party schedule for a 10-year-old with a superhero theme.
\end{itemize} \\ \hline
Write programs in the language of a machine's controller and store programs on media such as punch tapes, magnetic tapes, or disks. & Computer numerically controlled machine tool programmers, metal and plastic & 0.17 & 82 & Technical Assistance & \begin{itemize}
  \item Help me write g-code for drilling 5 holes evenly spaced along a 10 cm line.
\end{itemize} \\ \hline
Write articles, bulletins, sales letters, speeches, and other related informative, marketing and promotional material. & Copy writers & 0.17 & 83 & Generation & \begin{itemize}
  \item Write a sales letter for a new eco-friendly detergent.
\end{itemize} \\ \hline
Draw conclusions or make predictions based on data summaries or statistical analyses. & Biostatisticians & 0.17 & 84 & Summarization, Technical Assistance, Information Retrieval & \begin{itemize}
  \item Summarise key findings from this results section from clinical trial dataset.
  \item Write code in r that will run a logistic regression on this dataset to predict disease occurrence.
  \item What does a p-value of 0.01 mean in this context?
\end{itemize} \\ \hline
Assist students who need extra help with their coursework outside of class. & Biological science teachers, postsecondary & 0.17 & 85 & Reviewing Work, Information Retrieval & \begin{itemize}
  \item Check my lab report for accuracy
  \item Explain how photosynthesis works in 2–3 sentences.
\end{itemize} \\ \hline
Analyze business operations, trends, costs, revenues, financial commitments, and obligations to project future revenues and expenses or to provide advice. & Accountants & 0.17 & 86 & Reviewing Work, Data Structuring & \begin{itemize}
  \item Review my conclusions of this financial report and make sure they follow a logical conclusion
  \item Take my cells in in excel and convert the text in them into a format that is appropriate for a word document
\end{itemize} \\ \hline
Document findings of study and prepare recommendations for implementation of new systems, procedures, or organizational changes. & Management analysts & 0.16 & 87 & Summarization, Reviewing Work, Generation & \begin{itemize}
  \item Summarise the issues identified in this customer service audit.
  \item Evaluate this draft proposal for restructuring.
  \item Generate a list of actionable steps to improve efficiency.
\end{itemize} \\ \hline
Prepare reports, manuscripts, proposals, and technical manuals for use by other scientists and requestors, such as sponsors and customers. & Materials scientists & 0.16 & 88 & Generation, Reviewing Work & \begin{itemize}
  \item Draft the methods section of a technical report on alloy testing.
  \item Check this abstract for clarity and conciseness.
\end{itemize} \\ \hline
Evaluate sources of information to determine any limitations in terms of reliability or usability. & Statisticians & 0.16 & 89 & Reviewing Work, Information Retrieval & \begin{itemize}
  \item Evaluate the conclusion of this report based on the results section
  \item How do i determine the sample size required to measure a population?
\end{itemize} \\ \hline
Help programmers and systems analysts test and debug new programs. & Computer operators & 0.16 & 90 & Technical Assistance, Reviewing Work & \begin{itemize}
  \item Write a a program that will debug my Java code
  \item Review this error log and suggest fixes.
\end{itemize} \\ \hline
Provide assistance to students in college writing centers. & English language and literature teachers, postsecondary & 0.16 & 91 & Generation, Reviewing Work & \begin{itemize}
  \item Write a conclusion paragraph for an essay on climate change.
  \item Check this essay introduction for clarity and flow.
\end{itemize} \\ \hline
Answer customers' questions, and provide information on procedures or policies. & Cashiers & 0.16 & 92 & Information Retrieval & \begin{itemize}
  \item What's this store's return policy on electronics?
\end{itemize} \\ \hline
Create, analyze, report, convert, or transfer data, using specialized applications program software. & Geospatial information scientists and technologists & 0.16 & 93 & Data Structuring, Technical Assistance & \begin{itemize}
  \item Organise satellite imagery by resolution and date.
  \item Convert these shapefiles to geojson format
\end{itemize} \\ \hline
Develop briefings, brochures, multimedia presentations, web pages, promotional products, technical illustrations, and computer artwork for use in products, technical manuals, literature, newsletters, and slide shows. & Multimedia artists and animators & 0.16 & 94 & Generation & \begin{itemize}
  \item Create ideas for how i could design a visual diagram of a solar panel.
\end{itemize} \\ \hline
Read manuals, periodicals, and technical reports to learn how to develop programs that meet staff and user requirements. & Computer systems analysts & 0.16 & 95 & Summarization, Information Retrieval & \begin{itemize}
  \item Summarise this whitepaper on cloud architecture.
  \item What's the difference between rest and graphql apis?
\end{itemize} \\ \hline
Edit instructional materials, such as books, simulation exercises, lesson plans, instructor guides, and tests. & Instructional designers and technologists & 0.15 & 96 & Reviewing Work, Generation & \begin{itemize}
  \item Check this multiple-choice question for bias.
  \item Write an introduction to a module on computer literacy.
\end{itemize} \\ \hline
Write changes directly into compositions, or use computer software to make changes. & Music composers and arrangers & 0.15 & 97 & Generation, Reviewing Work & \begin{itemize}
  \item Add a bridge section to this melody in c major
  \item Review this transition between chorus and verse.
\end{itemize} \\ \hline
Design or develop software systems, using scientific analysis and mathematical models to predict and measure outcome and consequences of design. & Software developers, systems software & 0.15 & 98 & Information Retrieval, Technical Assistance & \begin{itemize}
  \item How can mathematical models assist in system development?
  \item Model a predictive maintenance system for industrial sensors.
\end{itemize} \\ \hline
Analyze test data, making computations as necessary, to determine test results. & Inspectors, testers, sorters, samplers, and weighers & 0.15 & 99 & Summarization, Technical Assistance & \begin{itemize}
  \item Summarise these qa results for executive reporting.
  \item Write code to calculate the mean and standard deviation for a dataset of product weights
\end{itemize} \\ \hline
Develop database architectural strategies at the modeling, design and implementation stages to address business or industry requirements. & Database architects & 0.14 & 100 & Technical Assistance & \begin{itemize}
  \item What's the best indexing strategy for a highly relational database and show me how to implement it
\end{itemize} \\ \hline
\end{longtable}

\end{document}